\documentclass[twoside]{article}

\usepackage{ifthen}
\usepackage{minitoc}

\usepackage[accepted]{aistats2021}

\usepackage{cancel}
\usepackage{nicefrac,xfrac}
\usepackage[round]{natbib}

\usepackage{xr}
\usepackage{hyperref}
\usepackage{booktabs}
\usepackage{mathtools}
\usepackage{color, xcolor}
\usepackage{latexsym}              
\usepackage{amsmath}               
\usepackage{amssymb}               
\usepackage{amsfonts}              
\usepackage{amsthm}                
\usepackage{multirow}
\usepackage{tikz}                  
\usetikzlibrary{arrows,positioning,shapes}
\usepackage{tcolorbox}
\usepackage{IEEEtrantools}
\usepackage{appendix}
\usepackage{algorithm}
\usepackage{tocloft}
\usepackage[noend]{algpseudocode}

\usepackage{multicol, blindtext}
\usepackage{subcaption}

\renewcommand{\thefootnote}{\fnsymbol{footnote}}
\newcommand*\samethanks[1][\value{footnote}]{\footnotemark[#1]}

\usepackage{xspace}
\usepackage[nolist,smaller]{acronym}
\definecolor{darkpink}{rgb}{0.91, 0.33, 0.5}
\definecolor{darkgreen}{rgb}{0.0, 0.5, 0.0}

\usepackage{bm}
\def\*#1{\bm{#1}}

\usepackage{ifthen}
\newif\ifapxinsamepage
\apxinsamepagetrue

\newcommand{\usec}{Sec.\xspace}
\newcommand{\utable}{Tab.\xspace}
\newcommand{\ueqn}{Eq.\xspace}
\newcommand{\uapendix}{App.\xspace}
\newcommand{\fig}{Fig.\xspace}
\newcommand{\acro}[1]{\textsc{#1}\xspace}
\newcommand{\GP}{\acro{\smaller GP}}
\newcommand{\titleGP}{\acro{GP}}
\newcommand{\VI}{\acro{\smaller VI}}
\newcommand{\BigGP}{{\Large $\mathcal{GP}$}}
\newcommand{\TGP}{\acro{\smaller TGP}}
\newcommand{\titleTGP}{\acro{TGP}}
\newcommand{\BATGP}{\acro{\smaller BA-TGP}}
\newcommand{\PETGP}{\acro{\smaller PE-TGP}}
\newcommand{\DGP}{\acro{\smaller DGP}}

\newcommand{\WGP}{\acro{\smaller WGP}}
\newcommand{\VWGP}{\acro{\smaller V-WGP}}
\newcommand{\titleVWGP}{\acro{V-WGP}}
\newcommand{\SAL}{\acro{\smaller SAL}}
\newcommand{\SALSP}{\acro{\smaller SAL+SP}}
\newcommand{\NN}{\acro{\smaller NN}}
\newcommand{\NNL}{\acro{\smaller NLL}}
\newcommand{\COV}{\acro{\smaller 95\% COVERAGE}}
\newcommand{\NLL}{\acro{\smaller NLL}}
\newcommand{\RMSE}{\acro{\smaller RMSE}}
\newcommand{\BNN}{\acro{\smaller BNN}}
\newcommand{\SVI}{\acro{\smaller SVI}}

\newcommand{\GSP}{\acro{\smaller G-SP}}
\newcommand{\titleGSP}{\acro{G-SP}}

\newcommand{\kk}{K}

\newcommand{\x}{\mathbf{x}} 
\newcommand{\X}{\mathbf{X}} 
\newcommand{\Xtest}{\mathbf{X^*}} 
\newcommand{\Y}{\mathbf{Y}} 
\newcommand{\Ytest}{\mathbf{Y^*}} 

\newcommand{\f}{\textbf{f}} 
\newcommand{\Tfinit}{\textbf{f}_{0}} 
\newcommand{\Tfinittest}{\f^*_{0}}   
\newcommand{\Tf}{\f_{\kk}}           
\newcommand{\Tftest}{\f^*_{\kk}}     
\newcommand{\Tfpos}[1]{\textbf{f}_{{#1}}} 

\newcommand{\J}[1]{\mathbf{J}_{#1}}

\newcommand{\T}{\mathbf{\mathbb{T}}} 

\newcommand{\G}{\mathbf{\mathbb{G}}} 
\newcommand{\arcsinh}{\text{arcsinh}} 
\newcommand{\priorjac}[1]{\left|\det \frac{\partial \G^{-1}_{\theta}(#1)}{\partial #1}\right| 
} 

\newcommand{\W}{\mathbf{W}} 

\newcommand{\E}{\mathbb{E}} 
\newcommand{\N}{\mathcal{N}} 
\newcommand{\LL}{\mathcal{L}} 
\newcommand{\TyL}{\bar{\LL}} 

\newcommand{\KL}{\acro{KL}} 
\newcommand{\KLD}{\acro{\smaller KL}} 
\newcommand{\ELBO}{\acro{\smaller ELBO}}
\newcommand{\ELL}{\acro{\smaller ELL}}
\newcommand{\SVGP}{\acro{\smaller SVGP}}

\newcommand{\uu}{\textbf{u}} 
\newcommand{\Tuinit}{\textbf{u}_{0}} 
\newcommand{\Tu}{\uu_{\kk}}           
\newcommand{\Tupos}[1]{\uu_{{#1}}} 

\newcommand{\Ty}{\Y_{\kk}}           
\newcommand{\Z}{\mathbf{Z}}           

\newcommand{\R}{\mathbb{R}}

\newcommand{\vm}{\mathbf{m}}
\newcommand{\MS}{\mathbf{S}}

\newcommand{\mbf}[1]{\mathbf{#1}}

\newcommand{\vu}{\mbf{u}}
\newcommand{\vf}{\mbf{f}}

\newcommand{\MY}{\mbf{Y}}

\newcommand{\BO}{\mathcal{O}}    

\DeclareMathOperator*{\argmin}{arg\,min}

\newcommand{\intg}[4]{\int_{#3}^{#4} #1 \, \mathrm{d}#2}

\newcommand{\Kxx}{K_{\X, \X}}
\newcommand{\Kxz}{K_{\X, \Z}}
\newcommand{\Kzx}{K_{\Z, \X}}

\newcommand{\iKzz}{K^{-1}_{\Z, \Z}}

\apxinsamepagetrue 

\ifapxinsamepage
{}
\else
{
\makeatletter
\newcommand*{\addFileDependency}[1]{
  \typeout{(#1)}
  \@addtofilelist{#1}
  \IfFileExists{#1}{}{\typeout{No file #1.}}
}
\makeatother

\newcommand*{\myexternaldocument}[1]{%
    \externaldocument{#1}%
    \addFileDependency{#1.tex}%
    \addFileDependency{#1.aux}%
}
\myexternaldocument{appendix}
}
\fi

\begin{document}

%

%
\runningauthor{Juan Maroñas, Oliver Hamelijnck, Jeremias Knoblauch, Theodoros Damoulas }

\twocolumn[

\aistatstitle{Transforming  Gaussian Processes With Normalizing Flows}

\aistatsauthor{ Juan Maroñas\samethanks[1]\samethanks[2] \And Oliver Hamelijnck\samethanks[2]  \And  Jeremias Knoblauch \And Theodoros Damoulas}

\aistatsaddress{ 
    PRHLT Research Center\\
    Universitat Politècnica \\
    de València  
\And 
    Dept. of CS  \\ 
    University of Warwick \\
    The Alan Turing Institute 
\And 
    Dept. of Statistics  \\ 
    University of Warwick \\
    The Alan Turing Institute 
\And
    Depts. of CS \& Statistics  \\ 
    University of Warwick \\
    The Alan Turing Institute 
} 
]

\begin{abstract}
  Gaussian Processes (\GP{}s) can be used as flexible, non-parametric function priors.
  Inspired by the growing body of work on Normalizing Flows, we enlarge this class of priors
  through a parametric invertible transformation that can be made input-dependent.
  Doing so also allows us to encode interpretable prior knowledge (e.g., boundedness constraints).
  We derive a variational approximation to the resulting Bayesian inference problem, which is as fast as stochastic variational \GP regression \citep{GPsBigData_hensman, dezfouli_black_box_gp:2015}.
  This makes the model a computationally efficient 
  alternative to other hierarchical extensions of \GP priors \citep{NIPS2012_4494,pmlr-v31-damianou13a}. 
  The resulting algorithm's computational and inferential  performance is excellent, 
  and we demonstrate this on a range of data sets.
  For example, even with only 5 inducing points and an input-dependent flow, our method is consistently competitive with a standard sparse \GP fitted using 100 inducing points.
\end{abstract}

\renewcommand{\thefootnote}{\arabic{footnote}}
\addtocontents{toc}{\protect\setcounter{tocdepth}{-1}}

\section{Introduction}

Gaussian Processes (\GP{}s) are perhaps the most well-known stochastic processes.
Their popularity derives from their two most important features:
not only are they infinite-dimensional generalizations of the multivariate normal distribution, but they also inherit numerous convenient properties from it.
Most importantly, like its finite-dimensional counterpart, the \GP is closed under marginalization and conditioning.
Together, these features
have made \GP{}s uniquely attractive for modeling natural phenomena in
molecular biology \citep{Einstein}, physics \citep{OUProcess}, and spatial statistics \citep{kriging}.

Within Machine Learning, \GP{}s are most commonly used as non-parametric Bayesian prior beliefs over functions, an idea dating back to \citet{OHagan} and significantly expanded by \citet{RasmussenGPsForML}.
Though \GP{} priors can describe many functions, an ongoing line of work has constructed ever more expressive function priors at the expense of computational complexity \citep{snelson_warpedgp,pmlr-v31-damianou13a,NIPS2010_4082,NIPS2012_4494,neuralprocesses}.
For instance, the work of \citet{pmlr-v31-damianou13a} and \citet{NIPS2012_4494} considers layered compositions of \GP{}s.
While these priors are more expressive than single \GP{}s, this comes at a price.
For example, in the Deep \GP (\DGP) inference algorithm of  \citet{salimbeni_dgp:2017}, computations are $\mathcal{O}(NM^2 \cdot K + M^3 \cdot K)$, where $N$ is the number of observations, $M<<N$ the number of inducing points and $K$ the total number of \GP{}s \citep[dozens per layer in the work of][]{salimbeni_dgp:2017}.

The current paper produces a method capable of largely eliminating this trade-off between the expressivity and computational complexity of function priors: We present a simple yet powerful way of enlargening the class of \GP priors without substantially increasing computational cost. 
Building on \citet{NIPS2010_4082} and \citet{NIPS2010_3996}, we apply parametric and invertible transformations (aka Normalizing Flows   \citep{pmlr-v37-rezende15}) to a \GP---yielding a Transformed \GP (\TGP). 
In contrast to previous approaches however, the \TGP also allows for Bayesian, input-dependent transformations.
The \TGP is more expressive than a \GP, and can encode additional prior knowledge about the latent function (e.g., boundedness constraints).
With a sparse variational inference scheme, the \TGP's run time is $\mathcal{O}(NM^2 + M^3)$---virtually identical to that of standard sparse variational \GP (\SVGP) regression \citep{GPsBigData_hensman}.
Further, the \TGP outperforms the \SVGP using only a fraction of its inducing points and produces test performances comparable to a multi-layer \DGP.

While this is not a focus of the current paper, our inference scheme can also easily incorporate transformations of the data.  
This means we also provide a faster approximation to a number of previous models, including the Warped Gaussian Process \citep{snelson_warpedgp,compositionally_warped_gp_rios:2019}.

\section{Motivation}
\label{sec:motivation}
Existing literature uses invertible transformations within \GP{}s either on the prior or the likelihood. %
For observations $(\X, \Y)$ and invertible mappings $\G, \T$, 
a generative model unifying both approaches is
\begin{IEEEeqnarray}{l}
    \begin{rcases}
          \Tfinit \sim \GP(\mu(\cdot), C(\cdot, \X)); \;\;
         \Tf = \G(\Tfinit)  \\
         \T(\Y) = \Tf + \epsilon; \:\:
         \epsilon \overset{iid}{\sim} \mathcal{N}(0, \Sigma).
    \end{rcases}
    \label{eqn:concept_eqn}
\end{IEEEeqnarray}

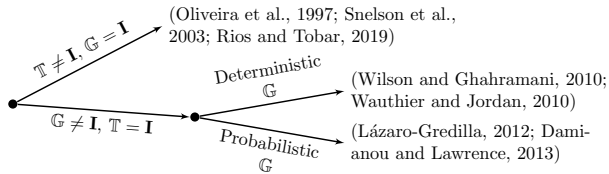
\begin{figure}[!t]

\centering

\resizebox{\columnwidth}{!}{

\begin{tikzpicture}[node distance=0.1cm]
\tikzstyle{line} = [draw, -latex'];
\tikzstyle{point}=[fill,circle,inner sep=0.5pt,minimum size=5pt];
\tikzstyle{write}=[rectangle,fill=white,text width=5cm];
\tikzstyle{write2}=[rectangle,fill=white,text width=6cm];

\node [point] (A) at  (-3.0,0.0) {};
\node [write2] (B1) at (3.0, 1.5) {\citep{oliveira_transformed_gaussians:1997,snelson_warpedgp,compositionally_warped_gp_rios:2019}};
\node [point] (B2) at  (0.5, -0.25) {};

\node [write] (C1) at (6.0, 0.25) {\citep{NIPS2010_4082,NIPS2010_3996}};
\node [write] (C2) at (6.0, -0.75) {\citep{NIPS2012_4494,pmlr-v31-damianou13a}};

\path [line,thick] (A) -- node [sloped, anchor=center, above, text width=2.0cm] {$\T \neq \mathbf{I}$, $\G = \mathbf{I}$} (B1.west);
\path[line,thick]  (A) -- node [sloped, anchor=center, below, text width=2.0cm] {$\G \neq \mathbf{I}$, $\T = \mathbf{I}$}(B2.west);

\path[line,thick]  (B2) -- node  [align=center,sloped, anchor=center, above, text width=2.0cm] {Deterministic $\G$}(C1.west);
\path[line,thick]  (B2) -- node  [align=center,sloped, anchor=center, below, text width=2.0cm] {Probabilistic $\G$}(C2.west);

\end{tikzpicture}
}
\caption{A simplified categorisation of some of the literature on transforming \GP{}s.}
\label{fig:prior_work}
\end{figure}

Denoting $\mathbf{I}$ as the identity function, this recovers standard \GP regression for $\G = \T = \mathbf{I}$.
Similarly, setting $\T \neq \mathbf{I}, \G = \mathbf{I}$ $/$ $\G \neq \mathbf{I}, \T = \mathbf{I}$  amounts to transforming only the likelihood $/$ prior.
Clearly, it is also possible to incorporate additional Bayesian priors about $\G$ and $\T$.
In this case, the transformation itself becomes probabilistic.
\fig \ref{fig:prior_work} illustrates this categorization.
\subsection{Related work}

The arguable more popular subcase of \ueqn\eqref{eqn:concept_eqn} is  applied to the likelihood ($\T \neq \mathbf{I}$, $\G = \mathbf{I}$).
The methods resulting from this strategy are well-studied in spatial statistics, and also known as Trans-kriging models \citep{diggle_model_geostatistics:2007}.
The earliest work is based on exponential transforms such as the Box-Cox \citep{box_transformations_analysis:1964}, but transformations soon took various other forms, including the Tukey transform \citep{tukey:1977}, hyperbolic transformations \citep{tsai_hyperbolic_normal:2017}, and the Sinh-Archsinh transforms \citep{sinh_arcsinh_transform_jones:2009}. 
Transformation parameters can be estimated \citep{snelson_warpedgp} or integrated out via Bayes' rule \citep{oliveira_transformed_gaussians:1997, mure_transgaussian_kriging:2018}. 
In essence, transforming the likelihood is an attempt at Gaussianization \citep{gaussianisation_chen:2000, gaussianization_flows_meng:2020}:
One hopes that $\T$ makes $\T(\Y) = \*Z$ into a standard \GP with additive noise \citep{transformed_additive_gps_lin:2019}.  
Note that whenever $\T$ is non-linear, this implies that $\Y = \T^{-1}(\*Z)$ is non-Gaussian with non-additive noise.
Thus, one can also see these transformations as aiming to fix model misspecification.
This means that unlike other approaches towards robustifying \GP{}s \citep[see e.g.][]{robustGPsStudentT1, robustGPsStudentT2,RDGP},
transformations of the likelihood make sense \textit{only} if one has sufficient domain knowledge to  locate the source of misspecification.

On the other hand, transformations of the prior ($\G \neq \textbf{I}, \T = \textbf{I}$) have \textit{no} implications for the likelihood model or error structure.
Further---and unlike transformations of the likelihoods---they are applicable for discrete-valued data, too.
This makes them more compatible with black box models
---and so more attractive to the Machine Learning community.
This does not mean that $\G$ cannot incorporate domain knowledge however---and we exploit this on two applications where we force the function prior to be non-negative.

These advantages have made the literature on transforming the \GP prior an active research area.
Some of its most prolific outcomes include \citet{NIPS2010_4082} as well as the work of \cite{NIPS2010_3996} and \cite{GPDensitySampler}. 
In all three papers, the parameterizations of the transforms are deterministic.
More recently, this was superseded by a probabilistic treatment \citep[e.g.][]{KarlasGPs}.
Deep \GP{}s \citep{ pmlr-v31-damianou13a} and Warped \GP{}s \citep{NIPS2012_4494} are perhaps the most prominent examples, and
transform a base \GP with a layered hierarchy of other \GP{}s. 
A different line of work transforms the prior via the input $\X$ \citep[see][]{manifold_gp_calandra:2016, deep_kernel_wilson:2016}, which induces non-stationarity relative to the original observation space without affecting the conditional Gaussianity of $\Y$.
\subsection{Computation}
Since the posterior of a standard \GP regression ($\G = \T = \mathbf{I}$) has closed form, it is important to determine how much $\G \neq \mathbf{I}$ or $\T \neq \mathbf{I}$ complicates computations.

When the likelihood is transformed ($\T \neq \mathbf{I}$, $\G = \mathbf{I}$), marginal likelihoods often have closed forms \citep[see e.g.][]{snelson_warpedgp}.
However, predictions need an explicit computation of the inverse $\T^{-1}$.

This leaves two options, both with considerable drawbacks: 
One can use approximation algorithms (e.g., Newton-Raphson) to approximate $\T^{-1}$, or one can constrain $\T$ to produce closed forms  \citep[see e.g.][]{compositionally_warped_gp_rios:2019}. 
The former bloats the computation and is sensitive to initial conditions, the latter constrains the model's flexibility. 

When the prior is transformed ($\T = \mathbf{I}$, $\G \neq \mathbf{I}$), the inverse $\G^{-1}$ does \textit{not} have to be computed explicitly. 
Such methods pose other challenges however: Their marginal likelihoods will not have closed form. 
In prior work, this has often resulted in rather coarse approximate inference.
For example, \citet{NIPS2010_4082} and  \cite{NIPS2010_3996} set $\G \neq \mathbf{I}$ to obtain non-Gaussian marginals, but are forced to use Laplace approximations for inference.

The problem is compounded if $\G$ itself is probabilistic, as is the case for Deep \GP{}s (\DGP{}s).
To address this issue, sparse \GP priors and Variational Inference (\VI) are typically used.
For instance, \citet{pmlr-v31-damianou13a, NIPS2012_4494} use a mean-field normal family. 
This is extended by \cite{salimbeni_dgp:2017} to capture uncertainty across layers.
Unfortunately, both approximations have drawbacks, leading to recent work advocating for structured variational families instead \citep{deep_gp_uncertainity:2020}.
This appears to produce better inferences, but also significantly increase the computational overhead.
\subsection{Our Contribution}
We design a Bayesian method that can match the performance of Deep \GP{}s at a fraction of the computational cost.
To achieve this, we consider Bayesian Neural Networks (\NN{}s) as input-dependent parametric transforms.
We then derive a sparse variational approximation extending the ideas of
\citet{VIinducingpoints_titsias, GPsBigData_hensman} and \citet{dezfouli_black_box_gp:2015}.

Our approximation is also the first scalable variational method for the methods of 
\cite{NIPS2010_4082, NIPS2010_3996}.
Further, our inference algorithm is applicable even if one also transforms the likelihood ($\G \neq \mathbf{I}, \T \neq \mathbf{I}$).
This means that it can easily be adapted to incorporate domain knowledge via $\G$ (and $\T$), and we show this on two examples.
\section{Model Description}

\usetikzlibrary{calc}
\begin{figure}[!t]

\centering

\resizebox{1.0\columnwidth}{!}{

\begin{tikzpicture}[node distance=2cm]

\tikzstyle{RV}=[shape=circle,line width=1.0pt,minimum size = 1.0cm, draw=black,fill=gray!20]
\tikzstyle{RVo}=[shape=circle,line width=1.0pt,minimum size = 1.0cm, draw=black,fill=white]
\tikzstyle{param}=[fill,circle,inner sep=1pt,minimum size=0.5pt]

\tikzstyle{box}=[rectangle,fill=none];

\node [RVo] (X) at (0.5,0.25) {$\X$};
\node [param, label = above:{ $\nu$}] (nu) at (0.5,1.0) {};

\node [RV,label={[xshift=-10pt,yshift=8pt]}] (f0) at (2.5,1.0) {$\Tfinit$};
\node [RV] (theta) at (2.5,-0.5) {$\theta$};

\node [RV,label={[xshift=-10pt,yshift=8pt]}] (fk) at (5.0,1.0) {$\Tf$};
\node [RV] (W) at (5.0,-0.5) {$\W$};

\node [RVo] (Y) at (7.0,1.0) {$\Y$};
\node [param, label = below:{ $\lambda$}] (lambda) at (7.0,-0.5) {};
\node[box] (G)  at (3.45,0.65) {{\tiny$\G_{\theta}(\Tfinit)$}};

\path [-,draw,thick] (G) -- ($ (f0) !.455! (fk) $);
\path [-,draw,thick] (G) -- ($ (theta) !0.63! (fk) $);

\draw[->,thick] (nu) edge (f0) ;
\draw[->,thick] (X) edge (f0) ;

\draw[->,thick] (lambda) edge (W) ;
\draw[->,thick] (X) edge (theta) ;

\draw[->,thick] (W) edge (theta) ;

\draw[->,thick] (f0) edge (fk) ;
\draw[->,thick] (theta) edge (fk) ;

\draw[->,thick] (fk) edge (Y) ;

\end{tikzpicture}}

\caption{
Plate diagram of the \TGP with Bayesian input-dependent flows. 
We use transformations  $\G_{\*\theta}$ with input-dependent (function-valued) parameters $\*\theta = \*\theta(\X, \W)$ to transform a base \GP $\Tfinit$ with kernel hyperparameters $\nu$ into a more expressive prior $\Tf$ about the functional relationship between $\X$ and $\Y$.
In practice, $\W$ will be weights of a Neural Network (\NN) with a Bayesian prior depending on hyperparameters $\lambda$.
%
%
}
\label{fig:input_dependent_graphical_model}

\end{figure}
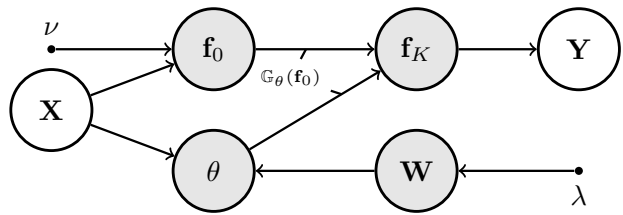
Given $N$ input-output tuples $\{(\X^{(i)},\Y^{(i)})\}^N_{i=1}$, we arrange them into matrices $\X\in \mathbb{R}^{N \times D_x}$ and $\Y \in \mathbb{R}^{N \times D_y}$. 
Throughout, our goal is the Bayesian learning over a set of functions $\mathcal{F} = \{f:\mathcal{X} \to \mathcal{Y}\}$.
To this end, we place a prior distribution over a subset of possible functions $\mathcal{F}$.
Given a $\GP$ specified via its mean and covariance functions $\mu(\cdot), C_{\nu}(\cdot, \cdot)$, 
we achieve this by additionally transforming it with $\kk$ invertible parametric transformations $\{\G_{{\theta}_k}\}_{k=0}^{K-1}$.
More precisely, we define for all $k=0,\dots K-1$ 
the functions $\G_{{\theta}_k}: \mathcal{F} \to \mathcal{F}$
as the individual transformations, $\G_{\*\theta} = \G_{{\theta}_0} \circ \G_{{\theta}_1} \circ \dots \circ \G_{{\theta}_{K-1}}$ as their composition and $\*\theta = \{\theta_0, \theta_1, \dots, \theta_{K-1}\}$ as the parameterization of this composition. 
Transformations of this kind have recently been popularized in a different context as flows \citep{pmlr-v37-rezende15}.
While our model applies for $\*\theta \in \mathbb{R}^p$, it also accommodates the case of function-valued (i.e. input-dependent) parameters $\*\theta: \mathcal{X} \to \mathbb{R}^p$ such as Neural Networks (\NN{}s).
\subsection{The Transformed Gaussian Process (\titleTGP)}

Taking $\Tfinit \sim \GP(\mu(\X), C_{\nu}(\X, \cdot))$ as a sample from the base \GP, we then define the \TGP as $\Tf = \G_{\*\theta}(\Tfinit)$.
For simplicity, the current paper restricts attention to element-wise mappings.
Because such mappings produce diagonal Jacobians, they only affect the marginals of the \GP, so that for any fixed $\X' \in \mathcal{X}$, $\Tf(\X') = \G_{\*\theta}(\Tfinit(\X'))$. Thus, we will often refer to them as \textit{diagonal}/\textit{marginal} transformations/flows.
Note that the resulting \TGP $\Tf$  can be seen as an input-dependent generalization of the Gaussian Copula Process discussed in \citet{NIPS2010_4082}.
\subsection{Input-dependent Flows}
A simple example for a marginal flow is given by stacking $K$ \SAL flows \citep{compositionally_warped_gp_rios:2019}:
\begin{IEEEeqnarray}{rCl}
\begin{rcases}
    \textbf{f}_{1} & = 
    d_1 \cdot \sinh (b_1\cdot\arcsinh(\Tfinit)-a_1) + c_1  \\
    & \dots  \\
    \Tf & = 
    d_\kk \cdot \sinh (b_\kk\cdot\arcsinh(\Tfpos{\kk-1})-a_\kk) + c_\kk 
\end{rcases}
\quad
\label{eq:Sal_flows}
\end{IEEEeqnarray}
In this example, $\G_{\*\theta}$ is not input-dependent and  $\theta_{j-1}=\{a_j,b_j,c_j,d_j\}$.
\fig\ref{fig:flow_plot} illustrates the effect of such a transform on a base \GP for $K=3$.
We could make the transformation input-dependent however, the only thing required is a reparameterization. In particular, one only has to replace the scalar parameters $a_j$, $b_j$, $c_j$, and $d_j$ with the function-valued parameters
$\alpha_j, \beta_j, \gamma_j, \delta_j:\mathcal{X} \to \mathbb{R}$.

We achieve this via Neural Networks
(\NN{}s) with $L$ layers so that for any fixed $\X' \in \mathcal{X}$, the transformation's parameters are $\{\alpha_j(\X'), \beta_j(\X'), \gamma_j(\X'), \delta_j(\X')\}_{j=0}^{K-1}$.
Thus, if the \NN's weights $\{\W^{l}\}_{l=1}^L$ are fitted without accounting for parameter uncertainty, $\*\theta = \{\W^{l}\}_{l=1}^L$.
Note that a model of this form will be able to model non-stationary processes.
We illustrate this using a range of warping functions at different locations in 
\uapendix B.7.2.
\subsection{Bayesian Priors on Flows}
However, we find that a Bayesian treatment of $\{\W^{l}\}_{l=1}^L$ significantly improves test set performance. 
This is hardly surprising; input-dependent flows in the form of \NN{}s introduce a considerable number of additional hyperparameters, making a naive implementation prone to over-fitting.
The reason for this is that enriching \GP priors with non-Bayesian flows provides additional fexibility via hyperparameters which are not regularized via a complexity penalty at inference time.
By placing a Bayesian prior $p(\W)$ on the network weights $\W = \{\W^l\}_{l=1}^L$, we effectively regularize the network weights and avoid this issue. 
This means that we integrate over $\{\W^{l}\}_{l=1}^L$, accounting for uncertainty in $\*\theta$.

Though the prior could be chosen arbitrarily, we consider the fully factorized normal prior $p_{\lambda}(\W) = \mathcal{N}(\W; 0, \lambda^{-1}I_{|\W|\times |\W|})$ throughout the paper.
The corresponding graphical model is given in \fig\ref{fig:input_dependent_graphical_model}, and the generative process is
\begin{IEEEeqnarray}{rCl}
    \Tfinit|\X &\sim &
    \GP(\mu(\X),C_{\nu}(\X,\cdot)) \hspace{0.6cm} \W \sim p_{\lambda}(\W)
    \nonumber \\
    \*\theta(\X, \W) & = & \NN(\X, \W) \hspace{0.6cm} \textbf{f}_K|\*\theta, \X, \W = \G_{{\*\theta}(\X, \W)}(\Tfinit)\nonumber 
    \nonumber
\end{IEEEeqnarray}
Unlike in previous work \citep[e.g.][]{NIPS2010_4082}, we not only quantify uncertainty about the parameter $\*\theta$, but also make it an input-dependent function.
\subsection{Induced Distributions}
\begin{figure}[!t]
    \centering
    \includegraphics[width=\columnwidth]{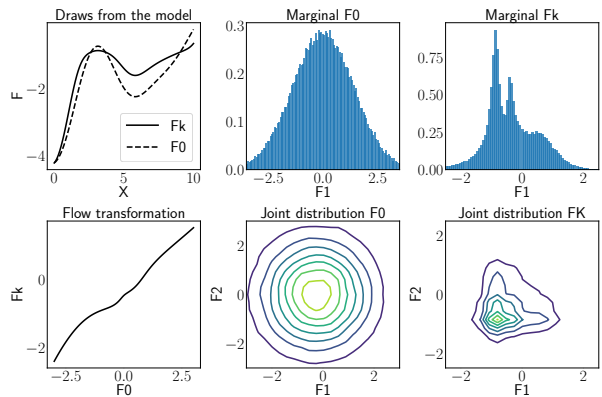}
    \caption{Flow constructed as in \ueqn\eqref{eq:Sal_flows} with $K=3$. The parameters of the flow were obtained from one of the experiments run in this work.}
    \label{fig:flow_plot}
\end{figure}

By virtue of an iterated application of the change of variable formula and the inverse function theorem \citep[see e.g.][]{pmlr-v37-rezende15}, the probability distributions induced by our transformations are: 
\begin{IEEEeqnarray}{rCl}
    p(\Tf|\G,\X) &= & p(\Tfinit|\X) \prod^{\kk-1}_{k=0}\left|\det\frac{\partial \G_{{\theta_k}}(\Tfpos{k})}{\partial \Tfpos{k}}\right|^{-1}.
    \nonumber
\end{IEEEeqnarray}
By using a marginal flow, Sklar's theorem \citep{Skla59} implies that the dependencies in $p(\Tfinit)$ and $p(\Tf)$ are driven by the same Copula---the \GP in our case.
Though the copula is the same, $\Tf$ will generally have non-Gaussian marginals (see \fig\ref{fig:flow_plot}).
While the current paper restricts attention to diagonal mappings for simplicity, the presented derivations and methods may be  extended
to non-diagonal transformations such as those in \citet{rios2020transport}. 
In practical terms, non-diagonal transformations could be used to model arbitrary copulas and correlation structures; and we elaborate on this version of the model in \uapendix \ref{mathematical_apx}. 
Whether $\G_{\*\theta}$ is a diagonal or non-diagonal transformation, we  require that the resulting $\Tf$ is a valid stochastic process (and thus a valid function prior). This amounts to checking whether the resulting collection of random variables satisfies the necessary consistency conditions, which holds by simple arguments for marginal flows  \citep[e.g.][]{rios2020transport}. 
In order to employ flows such as  Real NVP \citep{RealNVP}, one needs to prove that these conditions are still satisfied.
We leave this for future work, as the associated theory is highly dependent on the exact flow in question.

\section{Inference}
Performing inference for the Transformed Gaussian Process (\TGP) is generally intractable. 
Thus, we derive an efficient sparse variational approximation which is amenable to stochastic optimization \citep{GPsBigData_hensman}, utilizes inducing points through sparse \GP priors \citep{VIinducingpoints_titsias} and works with arbitrary likelihoods \citep{classificationsparseGPVIHensman,dezfouli_black_box_gp:2015}. 
An important part of this is a careful choice of the variational posterior, which eliminates the need to compute Jacobians or the inverse forms of $\G_{\*\theta}$ altogether, yielding a drastic speedup.
As a result, our inference algorithm is of order $\mathcal{O}(NM^2 + M^3)$ (for $M << N$) and parameters can be set via stochastic optimization.
This makes our method particularly suitable for large-scale applications---and much faster than previous approaches:
for example, the case of deterministic parametric transformations presented by \citet{NIPS2010_3996} and \citet{NIPS2010_4082} relies on $\mathcal{O}(N^3)$ Laplace approximations.  
Similarly, the approximations for the case of hierarchical probabilistic transformations presented in \citet{salimbeni_dgp:2017} are of order $\mathcal{O}(NM^2 \cdot K + M^3 \cdot K)$, where $K$ is the number of \GP{}s inside the Deep \GP \citep[dozens per layer in the work of][]{salimbeni_dgp:2017}.

\subsection{Sparse Variational Objective}

Variational methods frame inference as optimization by minimizing the Kullback-Leibler divergence (\KLD) between approximate and true posterior \citep{blei_vi:2017}.
It can also be interpreted as constrained finite-dimensional version of the infinite-dimensional variational problem characterizing the exact Bayesian posterior \citep{GVI}.
Rewriting this minimization as maximization, it becomes an Evidence Lower Bound (\ELBO) \citep{PMLRbishop}.

Sparse \GP{}s augment the prior with $M$ inducing points $\Tuinit \in \R^{M}$ at locations $\Z \in \R^{M \times D}$, typically with $M<N$.  These points act as `pseudo-observations' and allow low rank approximations to the \GP prior that circumvent the cubic costs traditionally associated with \GP inference \citep{unifying_sparse_gps:2005, williams_nystrom_gp:2000}.

Taking the inducing points into account, the sparsified and transformed prior of the \TGP is given by 
\begin{IEEEeqnarray}{rCl}
	p(\Tf, \Tu) = \underbrace{p(\Tfinit \mid \Tuinit)\J{\Tf}}_{=p(\Tf|\Tu)}\underbrace{p(\Tuinit)  \J{\Tu}}_{=p(\Tu)},
\label{eqn:sparse_prior}
\end{IEEEeqnarray}
where $p(\Tuinit)$ is a \GP prior of the same form as that for $\Tfinit$ in \ueqn\eqref{eqn:concept_eqn}
, $p(\Tfinit \mid \Tuinit)$ is conditionally Gaussian and $\J{\mathbf{a}} =  \prod^{\kk-1}_{k=0} \left|\det\frac{\partial \G_{{\theta_k}}(\mathbf{a})}{\mathbf{a}}\right|^{-1}$ is the (diagonal) Jacobian of the transformation of the stochastic process $\mathbf{a}$. \citet{rios2020transport} formally proves that stochastic processes transformed by such marginal flows induce valid stochastic processes. 
In turn, this guarantees that the transformed sparse stochastic process is consistent---and thus a valid function prior. 

One important property of the original bound proposed by \citet{VIinducingpoints_titsias} is that the variational posterior implicitly cancels the conditional, as this alleviates the need for computing $\mathcal{O}(N^3)$ matrix inverses \citep{SPGunderstanding}. 
Using the same algebraic tricks, we define our approximate posterior such that \textit{both} the conditional and the Jacobians cancel 
\begin{IEEEeqnarray}{rCl}
	q(\Tf, \Tu) = 
	p(\Tf|\Tu)
	\underbrace{
	    q(\Tuinit)\J{\Tu}
	}_{=q(\Tu)}
	\nonumber
\end{IEEEeqnarray}
where the $p(\Tf|\Tu)$ and $\J{\Tu}$ are defined as before 
and $q(\Tuinit)=\N(\Tuinit \mid \mathbf{m}, \mathbf{S})$ is a free form Gaussian with $\vm \in \R^{M \times 1}$ and $\MS \in \R^{M \times M}$. 
Another crucial side-effect of defining the variational posterior in this way is that it allows us to integrate out $\Tu$ analytically. 
Following \cite{GPsBigData_hensman} we do not collapse $q(\Tuinit)$ in order for stochastic Variational Inference to scale, resulting in the following \ELBO:
\begin{IEEEeqnarray}{rCl}
	\LL(\Y) = \E_{q(\Tf, \Tu)} \left[ \log \frac{   p(\Y \mid \Tfpos{\kk})\cancel{ p(\Tf \mid \Tu)}p(\Tuinit) \cancel{\J{\Tu}}}{\cancel{p(\Tf \mid \Tu)}q(\Tuinit)\cancel{\J{\Tu}}} \right]
	\nonumber
\end{IEEEeqnarray}
which  after some algebraic manipulations (see \uapendix \ref{mathematical_apx})
simplifies to our proposed variational lower bound:
\begin{IEEEeqnarray}{rCl}
	\sum^N_{n=1} \E_{q(\Tfpos{0,n})} \left[ \log p(\Y_n \mid \G_{\*\theta}(\Tfpos{0,n}))\right] - \KL \left[q(\Tuinit) \mid \mid p(\Tuinit) \right].
	\nonumber
\end{IEEEeqnarray}
The use of marginal flows and factorizing likelihoods results in the expected log likelihood (\ELL) term being decomposable across the latent variables $q(\Tfpos{0,n})$ and observations $\Y_n$---making it particularly suitable for stochastic variational inference and big $N$
\citep{GPsBigData_hensman}.
The individual \ELL components will generally be unavailable in closed form and computed using one-dimensional Gaussian quadrature \citep{classificationsparseGPVIHensman}.

\subsection{Justification of Approximate Family}
\label{sec:ablation_study}
To understand the effect of our proposed approximate family we now compare against the case where the variational posterior is constrained to be Gaussian. We denote this model as \GSP and provide the full derivation in \uapendix \ref{sec:appendix_gsp_derivation}. Because only the prior is transformed,  the  sparse conditional $p(\Tf \mid \Tu)$ no longer cancels; leading to an \ELBO requiring $\BO(N^3)$ computations and thus limiting $\GSP$ to only small scale experiments. Additionally, \GSP requires passing samples of the Gaussian approximate posterior through the inverse prior transformation which can be undefined, for example, with positive enforcing flows. For \GSP, we thus only consider flows that leave the output space unconstrained.

\begin{figure}[!t]
    \centering 
    \includegraphics[width=\columnwidth]{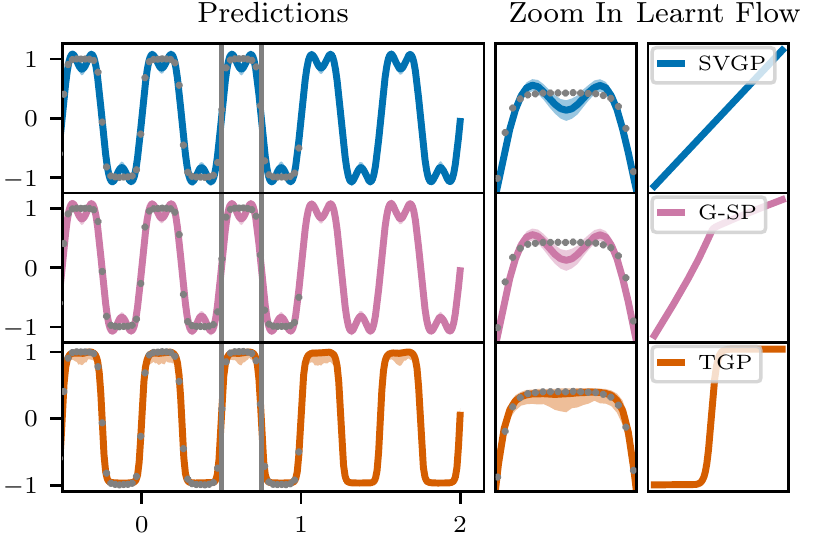}
    \caption{We generate 150 observations by passing evenly spaced evaluations of a sine curve, through a $\tanh$ flow and additionally applying Gaussian noise. For \TGP we use a $\tanh$ flow and for $\GSP$ we use a $\SAL$, because $\GSP$ requires a flow that does not constrain the output space. For all models we use a periodic kernel and perform a grid search across and a period initial value of $[0.6, 0.7, 0.8, 0.9, 1.0]$ and flow initialisation of identity, random and from data. We plot the best performing result from each model. The \SVGP is unable to accurately model the flat regions of the dataset, whereas \TGP can due to the prior transformation. \GSP uses a less suitable flow and so almost recovers the \SVGP.
    }
    \label{fig:ablation_study}
\end{figure}

As demonstrated in \fig \ref{fig:ablation_study} the \TGP is able to achieve a superior performance by fitting the flat regions of the data. The \GSP, although enriched with a flow on the prior, cannot learn more complex, space constraining flows, and instead learns a transformation that is close to identity and hence almost recovers the \SVGP. In comparison \TGP enriches both the prior and the approximate posterior family, increasing the flexibility of both whilst achieving superior performance and maintaining the computation benefits of the \SVGP.
\subsection{A Sparsification of Previous Models}
\label{sec:lik_transform}

Note that our approximation directly provides a new sparse variational inference algorithm for the models proposed by \cite{NIPS2010_4082} and \citet{NIPS2010_3996}.
In methodological terms, it is a direct generalization of \citet{GPsBigData_hensman} (for $\G_{\*\theta} \neq \mathbf{I}$).
Crucially, this means that while our model allows for substantially more expressive function priors, it inherits the computational efficiency of the standard variational \GP model.

While we focus mainly on prior transformations, our variational approximation also allows for transformations of the likelihood. Following \cite{snelson_warpedgp} the transformed likelihood is given by:
\begin{IEEEeqnarray}{rCl}
	p(\Y \mid \T, \Tf) = p(\T(\Y) \mid \Tf) \underbrace{\prod^{K-1}_{k=0}\left|\det\frac{\T_k(\Y_k)}{\Y_k}\right|}_{\J{\Ty}}.
	\nonumber
\end{IEEEeqnarray}
where the Jacobian term does not depend on $\Tfinit$. Substituting this into our bound results in 
\begin{IEEEeqnarray}{rCl}
	\TyL = \LL(\T(\Y)) +  \log \J{\Ty}.
	\nonumber
\end{IEEEeqnarray}
Setting $\G_{\*\theta} = \mathbf{I}$, this constitutes the first sparse variational approximation to the work of \citet{snelson_warpedgp} and \citet{compositionally_warped_gp_rios:2019}.
Unlike previous inference schemes available for these models, this makes them applicable to large-scale applications through sparse GPs and mini-batching.
We demonstrate this  contribution in our experiments, and call this new inference scheme variational Warped \GP (\VWGP).
\subsection{Bayesian Input-dependent Flows}
For input-dependent flows whose parameters have Bayesian priors, we specify priors and variational posteriors that are independent of $\Tf$.
For the case of the input-dependent flow being a \NN, this means that
\begin{IEEEeqnarray}{rCl}
\begin{aligned}
& p(\Tf,\W) = p(\Tf)p_{\lambda}(\W) \\
& q(\Tf,\W) = q(\Tf)q(\W)
\end{aligned}
\nonumber
\end{IEEEeqnarray}
Absorbing these additional terms into the \ELBO adds $-\KL[q(\W)||p(\W)]$ to our bound. Additionally, the \ELL term now requires integration over $q(\W)$.
The resulting integral could be approximated with $S$ Monte Carlo samples $\{W_s\}_{s=1}^S$ using reparameterization, allowing unbiased gradient estimates of low variance to be computed via the approximation  
\begin{IEEEeqnarray}{rCl}
\begin{aligned}
	&\sum^N_{n=1} \E_{q(\W)q(\Tfpos{0,n})} \left[ \log p(\Y_n \mid \G_{\*\theta}(\Tfpos{0,n}))\right] \approx\\   
	&\sum^N_{n=1}\frac{1}{S}\sum^S_{s=1} \E_{q(\Tfpos{0,n})} \left[ \log p(\Y_n \mid \G_{\*\theta_{(\X,\W_s)}}(\Tfpos{0,n}))\right]; 
\end{aligned}
	\nonumber
\end{IEEEeqnarray}
In our experiments however, we instead use the Monte Carlo Dropout approximation \citep{MCdropout} as it is more memory-efficient and avoids well-known pathologies introduced by mean field approximate posteriors $q(\W)$ \citep[e.g.][]{problemsVIturner}. 
To study the effect this has relative to standard Variational Bayes, we provide a comparative study in \uapendix{} \ref{apx:variational_bayes}.

\subsection{Prediction}

To predict using the \TGP ($\G_{\*\theta} \neq \mathbf{I}, \T = \mathbf{I}$) we substitute the true posterior for its approximation $q(\Tf)$. The predictive distribution is then given by
\begin{IEEEeqnarray}{rCl}
	p(\Y^*) = \int p(\Y^* \mid \G_{\*\theta}(\Tfinit) ) q(\Tfinit) \mathop{\mathrm{d}\Tfinit}.
	\nonumber
\end{IEEEeqnarray}
For predictions, we use one-dimensional quadrature to approximate the first moment of $p(\Y^*)$. 
Confidence intervals are obtained by sampling from $p(\Y^*)$. 
Further details on predicting with Bayesian input-dependent flows and transformed likelihoods are in \uapendix \ref{mathematical_apx}.

\begin{figure}[!t]
    \centering
    \includegraphics[]{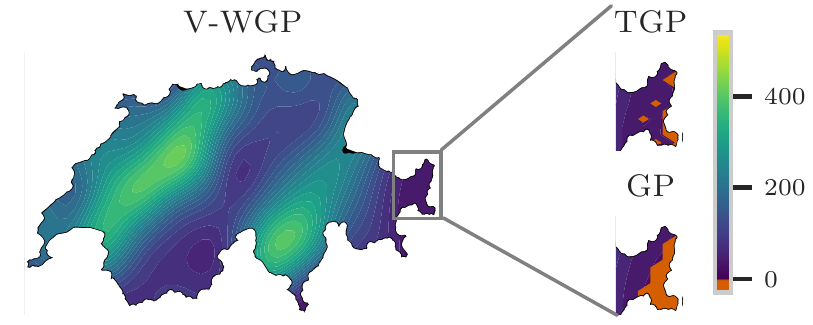}
\caption{Spatial median predictions from a \VWGP, \TGP and \GP on the Switzerland daily rainfall dataset (units in 10 $\mu m$). 
The \VWGP (left) is guaranteed to have non-negative predictions.
The \TGP and \GP (right)  do not and predict negative rainfall in Graub\"unden.}

\label{fig:rainfall_experiment}
\end{figure}
\section{Experiments}

We first compare the \TGP  with likelihood-transforming methods.
Second, we demonstrate the \TGP's excellent performance as a black box model by using input-dependent Bayesian flows on a range of UCI datasets \citep{UCI}. 
All code is written in PyTorch \citep{pytorch} using GPyTorch \citep{gardner_gpytorch:2018}. 
Details can be found in  \uapendix \ref{apx:additional_results}, and the code is publicly available at \texttt{https://github.com/jmaronas/TGP.pytorch}.

For all figures, we use the following acronyms: \TGP{} (non input-dependent \TGP{}), \BATGP (input-dependent \TGP{} with Bayesian flows) and \PETGP (input-dependent \TGP{} whose flow parameterization is obtained through a point estimate). 

\subsection{Applications}
\label{sec:applications}

We first study two applications to compare transformations of priors with those of likelihoods.
We restrict the \TGP to non-input-dependent flows and we compare it against a scalable variational approximation for the Warped \GP model of \citet{snelson_warpedgp} (henceforth \VWGP)  that we derive in \usec \ref{sec:lik_transform} and the Sparse Variational \GP (\SVGP) of \citet{GPsBigData_hensman}.

\begin{figure}[!t]
	\centering
    \includegraphics[]{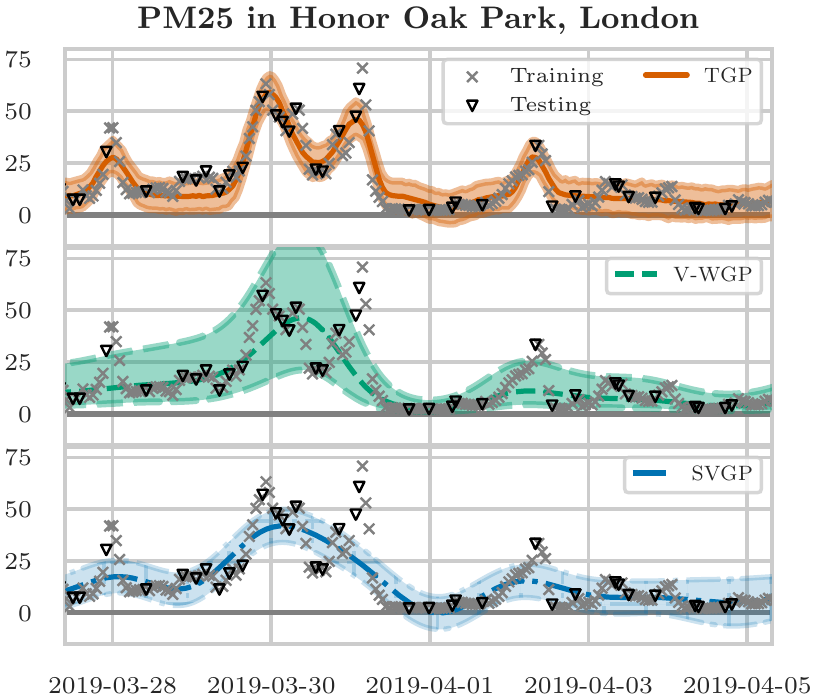}
\caption{ Model fits on PM25
with 5\% of inducing points.
The \TGP's added flexibility provides the best fit.
\textbf{Top}: the \TGP with compositional \SAL and softplus flow. \textbf{Middle}: \VWGP with the same flow,  but in reverse and applied to the likelihood. \textbf{Bottom}: \SVGP.  
}
\label{fig:aq_experiment_results}
\end{figure}

\textbf{Air Quality} \hspace{0.5cm} Consider Particulate Matter of 2.5 $\mu m$ size (PM25) in London \citep{laqn} as depicted in \fig\ref{fig:aq_experiment_results}. Measurements of PM25 are non-negative, exhibit periodic fluctuations due to vehicle traffic, and irregular peaks arising from weather conditions or traffic jams. Thus, we choose  a \SAL and softplus composition (\SALSP). This makes the \TGP's latent function positive and guarantees  $\T(\Y) \geq 0$ for the \VWGP.

\begin{figure}[!t]
    \begin{minipage}[b]{0.45\linewidth}
        \includegraphics[]{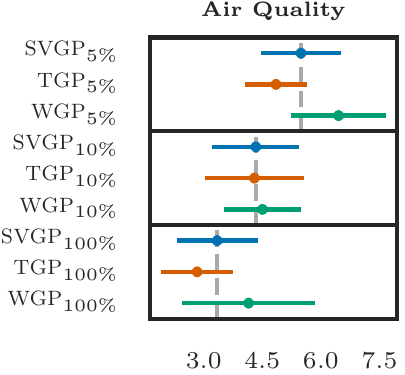}
    \end{minipage}
    \quad
    \begin{minipage}[b]{0.45\linewidth}
        \includegraphics[]{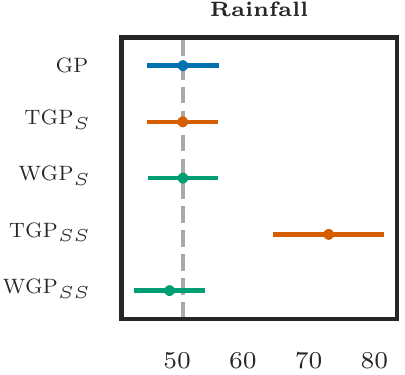}
    \end{minipage}
   \caption{\RMSE results (left is better) from the Air quality and Rainfall application. 
   \textbf{Left}:
   Air quality experiments with 5\%, 10\% and 100\% inducing points and a \SAL plus softplus flow. 
   The \TGP consistently outperforms the \GP and \VWGP because it can better fit to irregular patterns in the data.
   \textbf{Right}: 
   Rainfall experiments with 100\% inducing points and
   Softplus flows ($S$) versus \SAL plus softplus flows ($SS$). 
   When both the \TGP and \VWGP use the more expressive $SS$ flow, the $\VWGP$ is superior, reflecting that the source of misspecification is the likelihood, not the prior.}
\label{fig:aq_rainfall_results}
\end{figure}

The difference between methods is noticeable for low numbers of inducing points (see \fig \ref{fig:aq_experiment_results} \& \ref{fig:aq_rainfall_results}). 
As discussed in \usec \ref{sec:motivation}, the \VWGP implicitly models $\Y$ with non-additive noise while the \TGP transforms the prior, but models the noise additively.
Hence, the \TGP will attribute fluctuations to the underlying latent function, while the \VWGP is prone to absorb oscillations into the observation noise, as in \fig \ref{fig:aq_experiment_results}.
Unsurprisingly, the \TGP's fit is superior to that of the \GP due to its additional flexibility. However, even though $\G_{\*\theta}$ is chosen so that $\G_{\*\theta}(\Tfinit) \geq 0$, the \TGP assigns positive probability mass to PM25 being negative.

\textbf{Switzerland Rainfall} \hspace{0.5cm} We also model daily rainfall in Switzerland \citep{swiss_rainfall_dataset}, see \fig \ref{fig:rainfall_experiment}. As observations are non-negative, we again employ \SALSP flows.
Unlike the \TGP and \GP, the \VWGP does not fit the latent function to peaks in the data and guarantees positive predictions. The resulting smoother fit is desirable and explains why the \VWGP's predictive performance in \fig \ref{fig:aq_rainfall_results} outperforms that of the \TGP.

\subsection{Black Box results}

We also highlight our model's capability to learn arbitrary functions in a Bayesian way on a range of regression and classification problems.
Throughout, the \TGP uses 1- or 2-layer \NN{}s to parameterize input-dependent flows.
While the inverses of these flows would be difficult to approximate, inference for the \TGP can proceed without computing the inverse transformations.
This is a clear distinction to methods like the \VWGP, which relies on transforming the likelihood instead of the prior.
We present results for the negative log likelihood (\NLL) as they are representative for the overall findings and defer \RMSE values, \COV and accuracy metrics to \uapendix \ref{apx:additional_results} together with details on choosing flows and \NN architectures. 
For all plots, subscripts denote the number of inducing points used.
\begin{figure}[!t]
    \begin{subfigure}[t]{\columnwidth}
      \centering
        \includegraphics[width=\columnwidth]{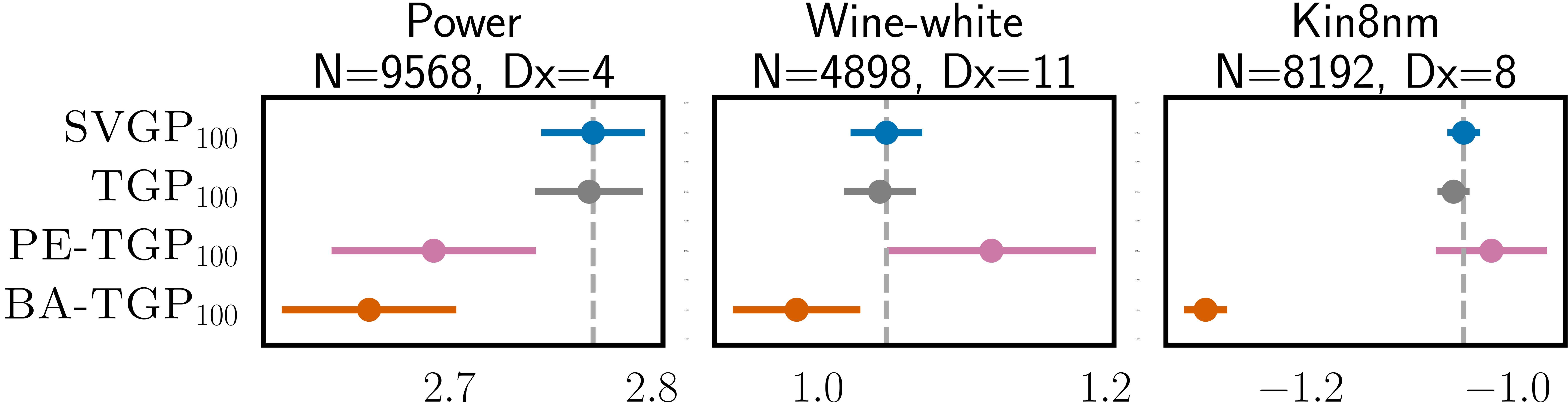}
    \end{subfigure}
    \begin{subfigure}[t]{\columnwidth}
      \centering
        \includegraphics[width=\columnwidth]{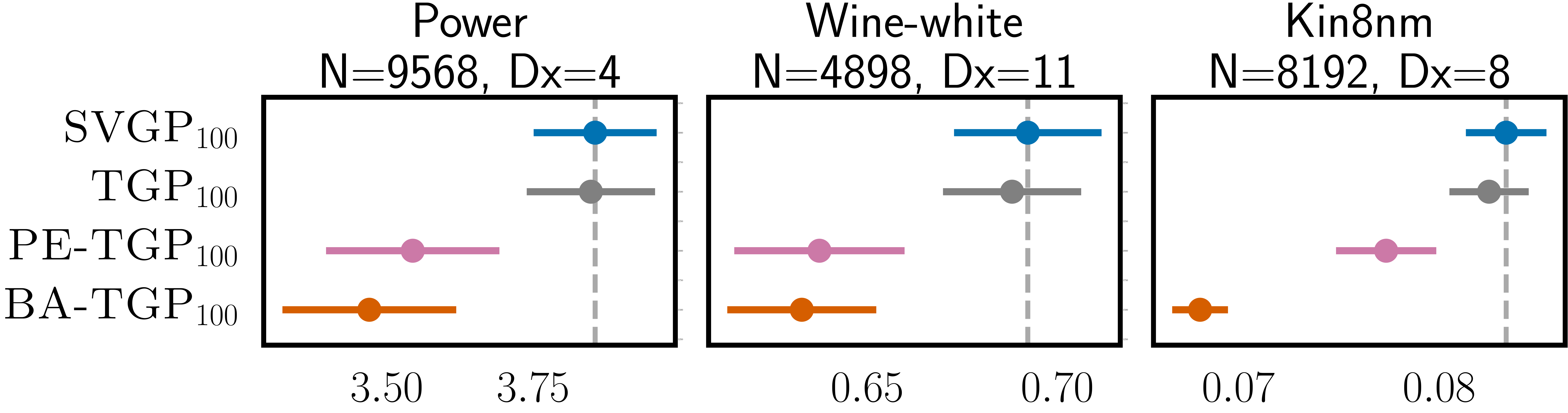}
    \end{subfigure}
     \caption{Comparison of \NNL (top; left is better) and \RMSE (bottom; left is better) for a standard \SVGP with a non input-dependent flow (\TGP),  the input-dependent counterpart indexed by a \NN when the \NN is fitted using a point estimate (\PETGP) or integrated out in a Bayesian fashion (\BATGP) }
     \label{fig:better_bayesian}
\end{figure}
\begin{figure}[!t]
    \centering
    \includegraphics[width=\columnwidth]{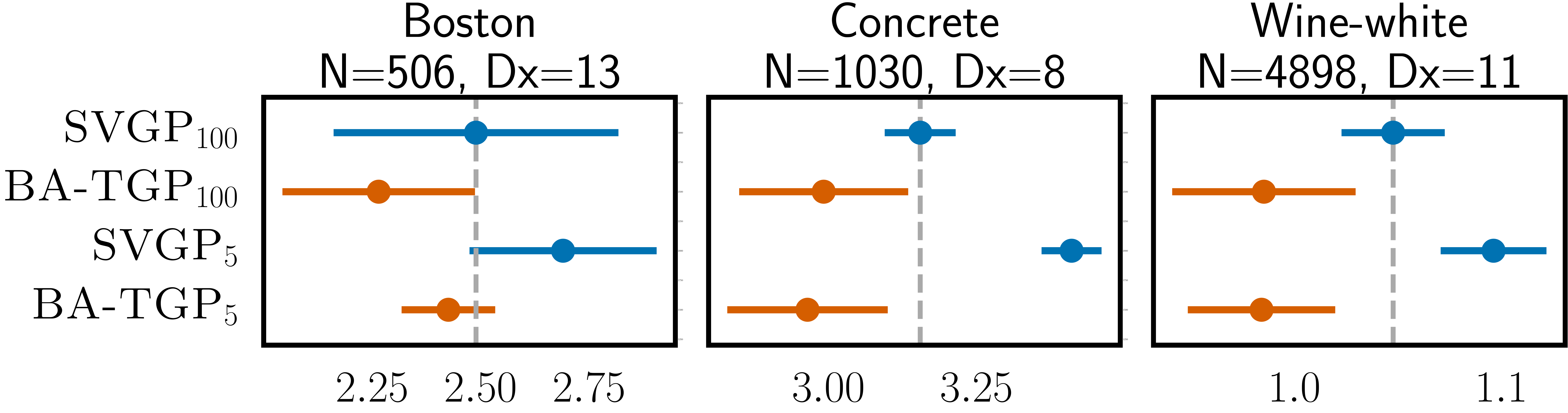}
    \caption{Comparing \NNL (left is better) for some medium-sized regressions with 5 and 100 inducing points. Remaining data sets in \uapendix \ref{apx:additional_results}.}
    \label{fig:5_inducing_points}
\end{figure}
\begin{figure}[!t]
\begin{subfigure}[t]{\columnwidth}
      \centering
        \includegraphics[width=\columnwidth]{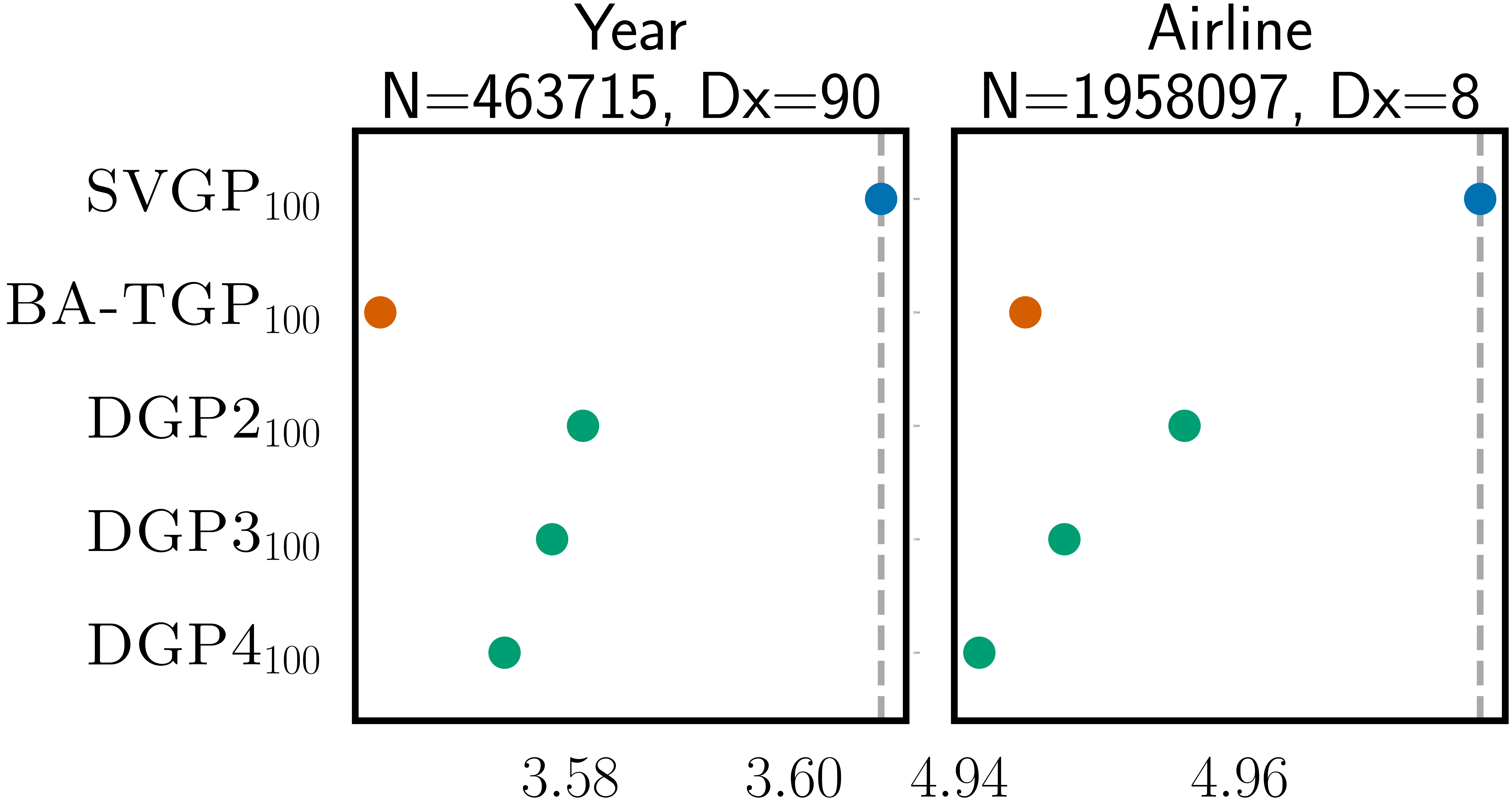}
    \end{subfigure}
     \caption{Comparing \NNL across 2 large data sets.}
     \label{fig:large_regressions}
\end{figure}
\begin{figure}[!t]
    \centering
    \includegraphics[width=\columnwidth]{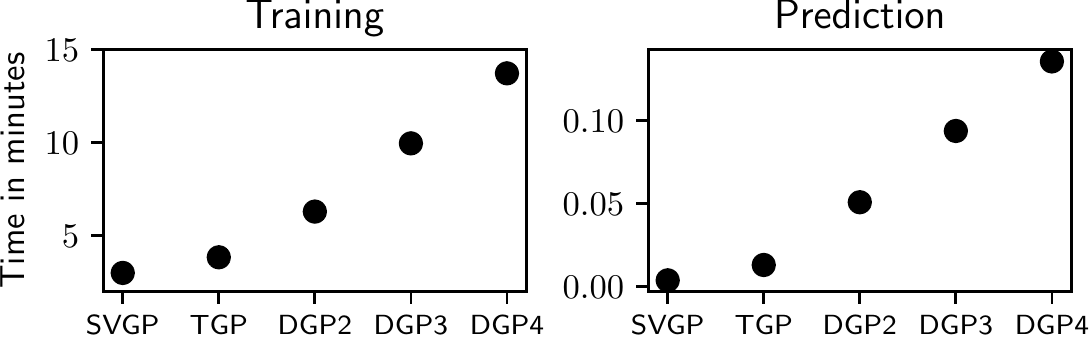}
    \caption{Average clock times for 100 runs with 1200 epochs on \texttt{energy}. 
    Predictions use 100 samples from the posterior.
    The variance of training and prediction repetitions is negligible ($<10^{-5}$).
    }
    \label{fig:computation_cost}
\end{figure}
\begin{figure*}[!t]
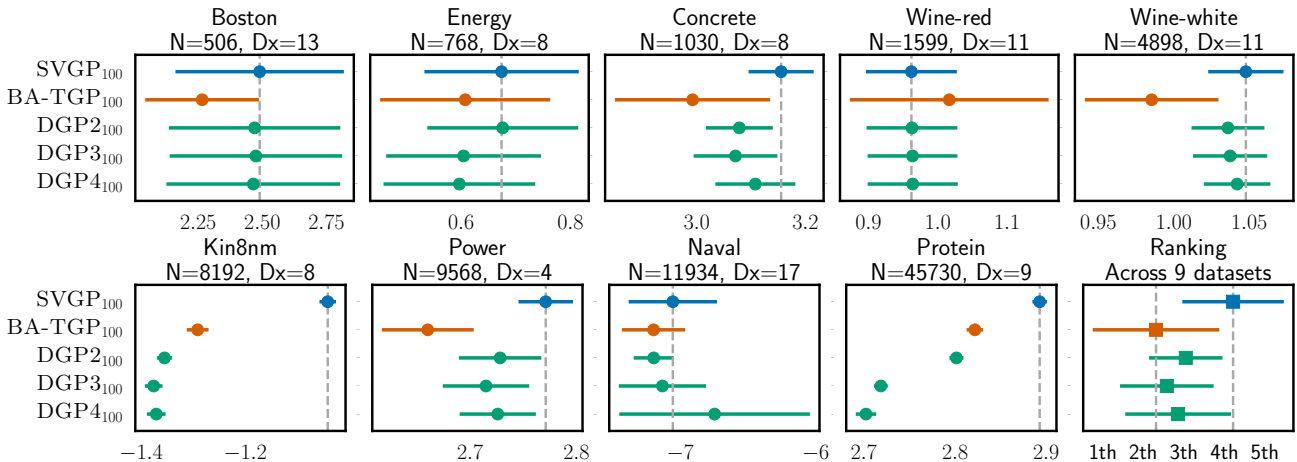

\begin{subfigure}[t]{\textwidth}
      \centering
       \includegraphics[width=\textwidth]{images/section5/medium_small_regression/LL_boston_energy_concrete_wine-red_wine-white_models_GP_BA_TGP_DGP_2_DGP_3_DGP_4_num_Z_100.pdf}  
\end{subfigure}\vspace{0.0cm}
\begin{subfigure}[t]{\textwidth}
      \centering
        \includegraphics[width=\textwidth]{images/section5/medium_small_regression/LL_kin8nm_power_naval_protein_ranking_models_GP_BA_TGP_DGP_2_DGP_3_DGP_4_num_Z_100.pdf}
    \end{subfigure}
     \caption{Comparing \NNL (left is better) across 9 data sets. \textbf{Bottom rightmost panel:} Ranking of the methods across all 9 data sets and repetitions shows that the \TGP performs as well as a 3- or 4-layer \DGP.}
     \label{fig:medium_small_regression}
\end{figure*}

\textbf{Bayesian Regularization} \hspace{0.5cm} First, we investigate the effect of Bayesian marginalization of the input-dependent flows  by comparing \RMSE and \NNL.
In particular, we use a \NN to induce input-dependence for the flow and compare the results obtained by using a point estimate via standard dropout \citep{standardDropout} versus those obtained with approximate Bayesian marginalization via Monte Carlo Dropout \citep{MCdropout}. 
The results are depicted in \fig \ref{fig:better_bayesian} and demonstrate that preventing the \NN from overfitting with a Bayesian treatment yields a significant performance boost in terms of predictive uncertainty. We also illustrate how non input-dependent flows are much less expressive than the input-dependent counterpart.

\textbf{Medium Scale Regression} \hspace{0.5cm} We compare \TGP{}s, \SVGP{}s and \DGP{}s with 2, 3, 4 layers on a range of medium-sized data sets.
For $D_x$ denoting the dimension of $\X$, each \DGP layer has at least $\text{min}(D_x,16)$ \GP's per layer, and we set kernel parameters and perform inference as in \citet{salimbeni_dgp:2017}.
Results over 10 training:test (90:10) splits are shown in \fig \ref{fig:medium_small_regression}. They demonstrate that the \TGP clearly outperforms the \GP and, on 5/9 datasets, the \TGP even manages to outperform the 4-layer \DGP.
When ranking the results, this implies an overall performance comparable to that of a 3-layer \DGP (see the rightmost bottom plot in \fig \ref{fig:medium_small_regression}). 

\textbf{Large Scale Regression} \hspace{0.5cm} We also benchmark the \TGP on two large scale regression datasets. The \texttt{Year} has 0.5M, and the \texttt{airline} around 2M training data points. 
\fig \ref{fig:large_regressions} shows the results and mirrors the findings for the medium-sized regressions.

\begin{figure}[!t]
    \centering
    \includegraphics[width=\columnwidth]{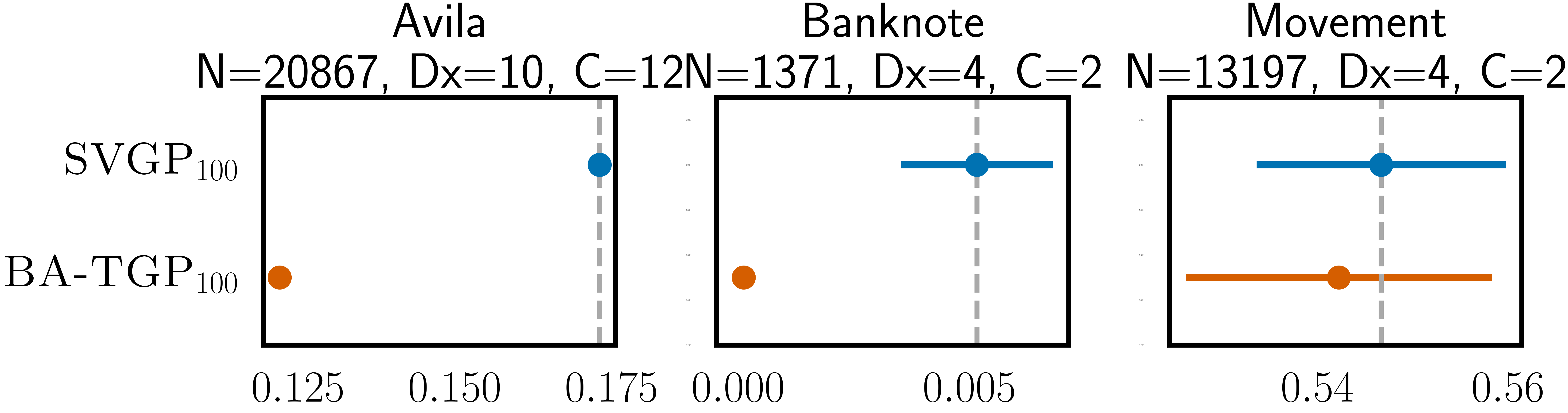}
    \caption{ Comparing \NNL (left is better) on three classification problems with up to $C=12$ classes.
    }
    \label{fig:classification}
\end{figure}

\textbf{Classification}\hspace{0.5cm}
Unlike transformations of the likelihood, prior transformations can be used to improve performance for discrete-valued data.
\fig\ref{fig:classification} makes this point using a range of classification problems.

\textbf{Computational Cost}\hspace{0.5cm} Impressively, the \TGP can match the 3-layer \DGP's performance at a fraction of the computational complexity: Even on the \texttt{energy} data set---where the \DGP has only 8 \GP{}s per layer---computation times for the 3-layer \DGP are 3$\times$ (training) and 10$\times$ (prediction) that of the \TGP (\fig\ref{fig:computation_cost}). 
This is true even though
our implementation of the \DGP fully exploits parallelization on a GPU cluster, while our current \TGP implementation does not exploit potential parallelization across \GP parameters. We further emphasize that our predictions use a 100 point quadrature integration rule per Monte Carlo sample, i.e we use $100\times100$ integration points, see \uapendix \ref{apx:predictions}, in contrast to the \DGP---which only uses 100. 

\textbf{Fewer Inducing Points} \hspace{0.5cm} The \TGP can match or outperform standard \GP{}s using 20 times fewer inducing points, and we illustrate this in \fig \ref{fig:5_inducing_points}. %
We provide evidence on additional data sets in \uapendix\ref{apx:additional_results} which also shows that $\text{Cov}[q(\Tfinit)]$ does not collapse to a point mass.
This implies that the Bayesian \NN flow does not make the \GP redundant---in fact, the \TGP's base \GP is essential to our model's quantification of uncertainty. 
\section{Conclusions and Future Work}

We introduced Transformed Gaussian Processes (\TGP{}s)---a new and flexible family of function priors---by enriching \GP{}s with parameterized, invertible, input-dependent, and Bayesian transformations (aka flows).
While the computational overhead of our inference scheme is comparable to that of sparse variational \GP regression, its predictive performance matches that of multi-layered \DGP{}s. 
The variational approximation  we derived also speeds up inference in  a host of other models \citep[e.g.][]{NIPS2010_4082, NIPS2010_3996, snelson_warpedgp} to $\mathcal{O}(NM^2 + M^3)$.
Our work can be used within inter-domain inducing point approximations \citep{interdomainGP_seminal,dutordoir2020sparse}, to improve multitask \GP{}s  \citep{bonilla_multi_task_gps:2007,review_multitask_GP, OlliesMTGP}, density estimation \citep{Conditional_density_estimationGP}, model calibration \citep{MARONAS2020194}, and probabilistic dimensionality reduction \citep{BGPLVM}. 

\ackaccepted{
J. M. is supported by grant FPI-UPV under grant agreement No 825111  DeepHealth Project, and by the Spanish National Ministry of Education through grant RTI2018-098091-B-I00.  O. H.,  J. K., and T. D. are funded by the Lloyd’s Register Foundation programme on Data Centric Engineering through the London Air Quality project. 
O. H. is funded through The Alan Turing Institute PhD fellowship programme. 
J. K. is funded by the EPSRC grant EP/L016710/1 as part of the Oxford-Warwick  Statistics Programme (OxWaSP) and the Facebook Fellowship Programme.
TD acknowledges support from UKRI Turing AI Fellowship EP/V02678X/1.
This work is also supported by EPSRC grant EP/T004134/1, by The Alan Turing Institute for Data Science and AI under EPSRC grant EP/N510129/1 in collaboration with the Greater London Authority, by the PRHLT Research Center and by the European Union under project Sistemas de frabricaci{\'o}n inteligentes para la industria 4.0 -- grant agreement IDIFEDER/2018/025. 

}

\bibliography{main}

 
\ifapxinsamepage
{

\appendix

\onecolumn
\aistatstitle{Appendix for `Transforming  Gaussian Processes With Normalizing Flows'}

\setlength{\cftbeforetoctitleskip}{-15em}
\tableofcontents



\clearpage
\addtocontents{toc}{\protect\setcounter{tocdepth}{3}}
\section{Mathematical Appendix}
\label{mathematical_apx}
In this appendix we provide all the mathematical details and derivations of the model presented in the main paper, plus some additional insights. We start with some definitions that will be used during the appendix. Then we move on the derivation of the $\ELBO$ and how we make predictions. The section is ended by strengthening the computational advantages implied by our definition. In order to highlight the role of each of the elements involved in our derivations, we provide a pictorial representation of our model in figure \fig \ref{fig:model_representation}.
\subsection{Definitions and Notation}
\paragraph{1.}  We first define some general notation. The finite subset of observations $\{\X^{(n)}\}^N_{n=1}$ and inducing points  $\{\Z^{(m)}\}^M_{m=1}$  from our stochastic process are stacked into matrices $\X$ and $\Z$ respectively. The corresponding function evaluations are stacked into vectors $\f$ and $\uu$. We denote with $\Tfinit$ and $\Tf$ to the function evaluations before and after applying the transformation $\G_{\theta}$. We denote specific locations $n$ or samples $s$ with additional subscripts e.g. $\Tfpos{0,n,s}$. 

\paragraph{2.} Given an invertible transformation $\G$, and the distribution $p(\Tf)$ induced by transforming samples from a base distribution $p(\Tfinit)$, then it holds that expectations  of any function $h()$ under $p(\Tf)$ can be computed by integrating w.r.t the base distribution $p(\Tfinit)$. This is formally known as \textit{probability under change of measure}. However throughout the document we will follow \cite{pmlr-v37-rezende15} and refer to it as \textit{LOTUS rule}. Formally, the above statement implies:
\begin{IEEEeqnarray}{rCl}
\E_{p(\Tf)}[h(\Tf)] = \E_{p(\Tfinit)}[h(\G_{\theta}(\Tfinit))]
\end{IEEEeqnarray}

\paragraph{3.}\label{definition3} For any transformation $\G$ that induces a valid stochastic process, see \cite{rios2020transport} for examples, it holds that the probability distribution at any finite subset of locations $\X$ and $\Z$ is given by:
\begin{IEEEeqnarray}{rCl}
    p(\Tf,\Tu) = p(\Tfinit,\Tuinit|\X,\Z)\prod^{\kk-1}_{k=0} \underbrace{\left|\det\left( \begin{matrix}\frac{\partial \G_{\theta}(\Tfpos{k})}{\partial \Tfpos{k}} & \frac{\partial \G_{\theta}(\Tfpos{k})}{\partial \Tupos{k}} \\ \frac{\partial \G_{\theta}(\Tupos{k})}{\partial \Tfpos{k}} & \frac{\partial \G_{\theta}(\Tupos{k})}{\partial \Tupos{k}}
    \end{matrix}\right)\right|^{-1}}_{\J{\Tfpos{k},\Tupos{k}}}
\end{IEEEeqnarray}
Where each element $\frac{\partial \G_{\theta}(\Tfpos{k})}{\partial \Tfpos{k}}$ is itself the Jacobian of the transformation of function evaluations at $\X$. By noting that the determinant of a block diagonal matrix can be computed as:
\begin{IEEEeqnarray}{rCl}
    \det\left(\begin{matrix}A & B \\ C & D\end{matrix}\right) = \det\left(A-BD^{-1}C \right) \det\left(D\right)
\end{IEEEeqnarray}
We can factorize the joint distribution $p(\Tf,\Tu)$ as follows:
\begin{IEEEeqnarray}{rCl}
\begin{aligned}
     p(\Tf,\Tu) & = p(\Tf|\Tu)p(\Tu)\\
     p(\Tf|\Tu) & = p(\Tfinit|\Tuinit) \prod^{\kk-1}_{k=0}\left|\det\left[\underbrace{\frac{\partial \G_{\theta}(\Tfpos{k})}{\partial \Tfpos{k}}}_{\J{\Tfpos{k}}}
     - \underbrace{\frac{\partial \G_{\theta}(\Tfpos{k})}{\partial \Tupos{k}}}_{\J{\Tfpos{k}|\Tupos{k}}}
     \left(\underbrace{\frac{\partial \G_{\theta}(\Tupos{k})}{\partial \Tupos{k}}}_{\J{\Tupos{k}}}\right)^{-1}
     \underbrace{\frac{\partial \G_{\theta}(\Tupos{k})}{\partial \Tfpos{k}}}_{\J{\Tupos{k}|\Tfpos{k}}}\right]\right|^{-1}\nonumber\\
     p(\Tu)      & = p(\Tuinit) \prod^{\kk-1}_{k=0}\left|\det\frac{\partial \G_{\theta}(\Tupos{k})}{\partial \Tupos{k}}\right| ^{-1}
\end{aligned}\label{apx_eqn:joint_prior}
\end{IEEEeqnarray}
where we make use of  $p(\Tf|\Tu)=\nicefrac{p(\Tf,\Tu)}{p(\Tu)}$ to derive the expression for the conditional distribution. Furthermore, note that for marginal flows, $B$ and $C$  are evaluated to zero and we recover the distributions used throughout the inference section in the main paper: 
\begin{IEEEeqnarray}{rCl}
 \begin{aligned}
     p(\Tf,\Tu) & = p(\Tf|\Tu)p(\Tu)\label{apx_eqn:joint_prior_diagonal}\\
     p(\Tf|\Tu) & = p(\Tfinit|\Tuinit)  \prod^{\kk-1}_{k=0}\left|\det\frac{\partial \G_{\theta}(\Tfpos{k})}{\partial \Tfpos{k}}\right|^{-1}\\
     p(\Tu)      & = p(\Tuinit) \prod^{\kk-1}_{k=0}\left|\det\frac{\partial \G_{\theta}(\Tupos{k})}{\partial \Tupos{k}}\right|^{-1}
 \end{aligned} \nonumber
\end{IEEEeqnarray}

where now  $\frac{\partial \G_{\theta}(\Tfpos{k})}{\partial \Tfpos{k}}$ is a diagonal matrix where the elements of the diagonal are given by $\frac{\partial \G_{\theta}(\Tfpos{k,n})}{\partial \Tfpos{k,n}}$.
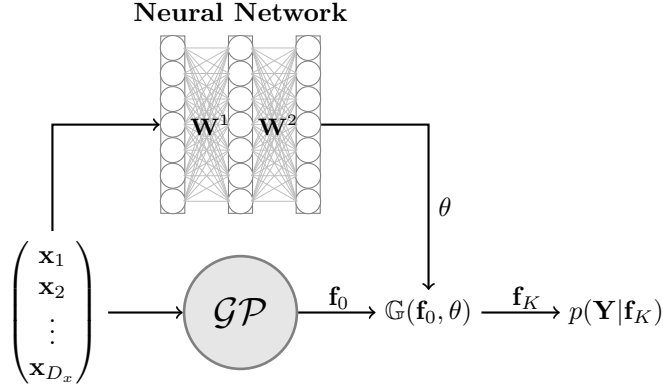
\begin{figure}[!t]

\centering

\begin{tikzpicture}[node distance=2cm]
\tikzstyle{GP}=[shape=circle,line width=1.0pt,minimum width=1.5cm,draw=black!50,fill=black!10]
\tikzstyle{Nnet}=[shape=rectangle,minimum height=2.36cm, minimum width=0.32cm,draw=black!50,fill=white!10]
\tikzstyle{neuron}=[shape=circle,minimum width=0.3cm,draw=black!50,fill=white!10]
\tikzstyle{dots}=[shape=circle]
\tikzstyle{write}=[]

\node [write] (x_concat) at (-1.0,1.0) {$\begin{pmatrix}\x_1\\\x_2\\\vdots\\\x_{D_x}\end{pmatrix}$};

\node [GP] (gp) at (1.5, 1.0) {\BigGP};

\node [Nnet]   (nnet)          at (0.6,3.5)  {};
\node [neuron] (n11)           at (0.6,4.52) {};
\node [neuron] (n12)           at (0.6,4.18) {};
\node [neuron] (n13)           at (0.6,3.84) {};
\node [neuron] (n14_middle)    at (0.6,3.5)  {};
\node [neuron] (n14)           at (0.6,3.16) {};
\node [neuron] (n15)           at (0.6,2.82) {};
\node [neuron] (n16)           at (0.6,2.48) {};

\node [write]  (nnet_name) at (1.5,5.0)  {\textbf{Neural Network}};
\node [Nnet]   (nnet) at (1.5,3.5)  {};
\node [neuron] (n21)   at (1.5,4.52) {};
\node [neuron] (n22)   at (1.5,4.18) {};
\node [neuron] (n23)   at (1.5,3.84) {};
\node [neuron] (n24)   at (1.5,3.5)  {};
\node [neuron] (n25)   at (1.5,3.16) {};
\node [neuron] (n26)   at (1.5,2.82) {};
\node [neuron] (n27)   at (1.5,2.48) {};

\node [Nnet]   (nnet) at (2.4,3.5)  {};
\node [neuron] (n1)   at (2.4,4.52) {};
\node [neuron] (n1)   at (2.4,4.18) {};
\node [neuron] (n1)   at (2.4,3.84) {};
\node [neuron] (n_middle_output)   at (2.4,3.5)  {};
\node [neuron] (n1)   at (2.4,3.16) {};
\node [neuron] (n1)   at (2.4,2.82) {};
\node [neuron] (n1)   at (2.4,2.48) {};

\draw[->,thick] (x_concat) edge (gp);
\draw[->,thick] (x_concat) -- (-1.0,3.5) -- (n14_middle);

\foreach \x in {0,...,6}
\foreach \y in {0,...,6}
   \draw[gray!50,line width=0mm] (0.75,4.52-\y*0.34) -- (1.35,4.52-\x*0.34);
\foreach \x in {0,...,6}
\foreach \y in {0,...,6}
   \draw[gray!50,line width=0mm] (1.65,4.52-\y*0.34) -- (2.25,4.52-\x*0.34); 

\node [write]  (layer1_weight) at (1.1,3.5) {{\small $ \W^{1}$}};
\node [write]  (layer2_weight) at (2.0,3.5) {{\small $ \W^{2}$}};

\node [write] (flow) at (4.0, 1.0)  {$\G(\Tfinit,\theta)$};
\node [write] (samplef0) at (2.75,1.2) {{ $ \Tfinit $ }};
\node [write] (samplefk) at (5.25,1.2) {{ $ \Tf $ }};

\node [write] (likelihood) at (6.5, 1.0)  {$p(\Y|\Tf)$};

\draw[->,thick] (gp) edge (flow) ;

\draw[->,thick] (n_middle_output)  -- (4.0,3.5) --  node[right] {$\theta$} (flow) ;

\draw[->,thick] (flow) edge (likelihood) ;

\end{tikzpicture}

\caption{A pictorial representation of our general formulation that highlights the role of the Neural Network. As seen in the figure, the output of the neural network gives the parameters of the flow $\G$. We can further incorporate uncertainty into this parameters by defining a prior $p(\W)$ over the \NN parameters. See the graphical model in the main article and the details in this appendix.}
\label{fig:model_representation}
\end{figure}

\subsection{Variational Lower Bound Derivation }
We now provide a detailed derivation of the variational lower bound of \TGP for both the marginal and non-marginal case. 
In order to reduce the computational cost of directly approximating the posterior at training locations $\X$ \citep{OpperVIGP}, we make use of a sparse approximation. 
We begin by first defining the sparse prior and approximate variational posterior and then derive the respective lower bounds for marginal and non-marginal flows.
As we will see the only difference between marginal and non marginal flows is that in the later the inducing points cannot be integrated with analytically, and so we resort to a Monte Carlo approximation.

\subsubsection{Sparse Prior}
For computational efficiency we follow \citet{VIinducingpoints_titsias, GPsBigData_hensman} and augment the \TGP prior  with inducing points $\Tu$ at inducing locations $\Z$. The sparse prior of \TGP is defined by:
\begin{equation}
	p(\Tf, \Tu) = p(\Tf \mid \Tu) p(\Tu) = p(\Tfinit \mid \Tuinit) p(\Tuinit) \J{\Tf, \Tu}
\label{eqn:app_sparse_prior}
\end{equation}
where the $p(\Tuinit)$ is a GP and the conditional $p(\Tfinit \mid \Tuinit)$ is a standard Gaussian conditional:
\begin{equation}
\begin{aligned}
	p(\Tuinit) &= \N(\Tuinit \mid 0, K_{\Z, \Z}) \\
	p(\Tfinit \mid \Tuinit) &= \N(\Tfinit \mid K_{\X, \Z} K^{-1}_{\Z, \Z} \Tuinit, K_{\X, \X} - K_{\X, \Z} K^{-1}_{\Z, \Z} K_{\Z, \X}) 
\end{aligned}
\label{eqn:appendix_sparse_prior_def}
\end{equation}
and $K$ is positive-semi definite kernel function. As described in the main paper, to sparsify this prior the transformation $\G$ must induce a valid stochastic process. It holds that marginal flows always induce a valid stochastic process, but some care has to be taken in the definition of non-marginal flows. We refer the reader to \citep{rios2020transport} for examples of non-marginal transformations $\G$ that transforms $p(\Tfinit)$ in a consistent way.

\subsubsection{Approximate posterior}
Following \citet{VIinducingpoints_titsias} we define our approximate posterior such that the conditional terms cancel:
\begin{equation}
	q(\Tf \mid \Tu) = p(\Tf \mid \Tu) q(\Tu)
\end{equation}
where 
\begin{equation}
	q(\Tu) = \N(\Tuinit \mid \vm, \MS) \J{\Tu}
\end{equation}
with $\vm \in \mathbb{R}^{M \times 1}$ and $\MS \in \mathbb{R}^{M \times M}$ being the variational parameters and $p(\Tf|\Tu)$ is given by definition 3.

\subsubsection{ELBO}

Following a similar derivation as \cite{GPsBigData_hensman} \& \cite{blei_vi:2017}, we can arrive at the following \ELBO:
\begin{equation}
\begin{aligned}
	\ELBO &= \E_{q(\Tf,\Tu)}  \left[ \log \frac{\prod_n p(\Y_n \mid \Tfpos{\kk,n}) p(\Tf,\Tu)}{q(\Tf,\Tu)} \right] \\
	&= \underbrace{\sum^N_n \E_{q(\Tfpos{\kk}, \Tu)}  \left[ \log p(\Y_n \mid \Tfpos{\kk,n}) \right]}_{\ELL} + \underbrace{\E_{q(\Tf,\Tu)}  \left[ \log \frac{p(\Tf,\Tu)}{q(\Tf,\Tu)} \right]}_{-\KLD} 
\end{aligned}
\label{apx_eqn:elbo}
\end{equation}
In order to highlight the difference between using marginal and non-marginal flows, we break the derivation of the final lower bound into the $\KLD$ and the expected log likelihood (\ELL). As we will illustrate, only the \ELL depends on the type of transformation (marginal or non) and so we first deal with the (negative) \KL term.

\subsubsection{KL divergence}

It turns out that the $\KLD$ between the prior and the approximate posterior on the transformed space $\Tf$, is given by the same $\KLD$ on the original space $\Tfinit$:
\begin{equation}
\begin{aligned}
	\KL &= - \E_{q(\Tf,\Tu)}  \left[ \log \frac{\cancel{p(\Tf \mid \Tu)} p(\Tuinit) \cancel{\J{\Tu}} }{\cancel{p(\Tf \mid \Tu)} q(\Tuinit) \cancel{\J{\Tu}}} \right] \\
	 &= - \E_{q(\Tu)}  \left[ \log \frac{ p(\Tuinit) }{ q(\Tuinit)} \right] \\
	&=\underbrace{-\E_{q(\Tuinit)}  \left[ \log \frac{p(\Tuinit)}{q(\Tuinit)} \right]}_{\coloneqq\KL[q(\Tuinit)||p(\Tuinit)]}
\end{aligned}
\end{equation}
where we have first marginalized $\Tf$ and then applied the LOTUS rule. By defining the approximate posterior as being transformed by the same flow as the prior it allows both the Jacobian and conditional distribution $p(\Tf|\Tu)$ to cancel. This not only alleviates costly $\mathcal{O}(N^3)$ computations coming from the \KLD between the conditional distributions, but also simplifies the \KLD to simply be between two Gaussian distributions.

\subsubsection{ ELL with Marginal Flows} 
We now derive the expected log term for the case in which $\G$ is a marginal flow. To do so we first derive the form of $q(\Tf)$ by analytically integrating out the inducing points $\Tu$. We then marginalize $\Tf$ such that the \ELL term decomposes into components of $\Tfpos{\kk, n}$ and $\Y_n$. The marginal $q(\Tf)$ is given by:
\begin{IEEEeqnarray}{rCl}
\begin{aligned}
        & q(\Tf) = \int q(\Tf,\Tu)d\Tu \\
        & = \int p(\Tfinit|\Tuinit) q(\Tuinit) \J{\Tf,\Tu} d\Tu
\end{aligned}
\end{IEEEeqnarray}

Because $\G$ is a marginal flow the Jacobian matrix is diagonal and decomposes such that $\J{\Tf,\Tu}=\J{\Tf}\J{\Tu}$. Substituting this into the above and applying LOTUS rule by recognizing this as an expectation w.r.t $q(\Tu)$:
\begin{IEEEeqnarray}{rCl}
\begin{aligned}
     q(\Tf) & =  \int p(\Tfinit|\Tuinit) q(\Tuinit) \J{\Tf}\J{\Tu} d\Tu \\
        & = \J{\Tf} \int p(\Tfinit|\Tuinit)q(\Tuinit) d\Tuinit  \\
        &  = \J{\Tf} q(\Tfinit)
\end{aligned}
\end{IEEEeqnarray}
where $\J{\Tf}$ does not depend on $\Tuinit$ and the marginal $q(\Tfinit)$ is:
\begin{IEEEeqnarray}{rCl}
\begin{aligned}
    q(\Tfinit) = \N(\Tfinit \mid K_{\X, \Z} K^{-1}_{\Z, \Z} \vm, K_{\X, \X} - K_{\X, \Z} K^{-1}_{\Z, \Z} \left[K_{\Z, \Z} + \MS \right] K^{-1}_{\Z, \Z} K_{\Z, \X})
\end{aligned}
\end{IEEEeqnarray}
Making use of this derivation of $q(\Tf)$ we can simplify the \ELL term in \ueqn\ref{apx_eqn:elbo}:%
\begin{IEEEeqnarray}{rCl}
\begin{aligned}
    \ELL &= \sum^N_n \E_{q(\Tfpos{\kk}, \Tu)}  \left[ \log p(\Y_n \mid \Tfpos{\kk,n}) \right]]\\
    & = \sum_n\E_{q(\Tfinit)} \left[ \log p(\Y_n \mid \Tfpos{\kk,n})\right] \\
    & = \sum_n\E_{q(\Tfpos{0,n})}\left[ \log p(\Y_n \mid \Tfpos{\kk,n})\right],
\end{aligned}
\end{IEEEeqnarray}

where we have now applied LOTUS rule over the expectation w.r.t $q(\Tf)$ after integrating out $\Tu$. We finally integrate out all the elements $\Tfinit$ but $\Tfpos{0,n}$ from our variational posterior by noting that $\Y_n$ only depends on the function evaluation at position $n$. Thus, the \ELL term now factorizes across $\Y_n$ and the latent variables $\Tfpos{0, n}$, as required for \SVI. Plugging this into the \ELBO:
\begin{equation}
	\ELBO = \sum^N_n \E_{q(\Tfpos{0, n})}  \left[ \log p(\Y_n \mid \G(\Tfpos{0,n})) \right] + \E_{q(\Tuinit)}  \left[ \log \frac{p(\Tuinit)}{q(\Tuinit)} \right]
\label{apx_eqn:elbo_marginal}
\end{equation}
recovers the lower bound presented in the main paper. 
\subsubsection{ ELL with Non Marginal Flows}\label{sec:appendix_non_marginal_flows}
We now present a generalization of the presented inference algorithm to include non-marginal flows where, as before, we only require that $\G$ induces a valid stochastic process. The key difference between using marginal and non marginal flows is that for non-marginal flows we will not be able to, in general, analytically integrate out the inducing points $\Tu$. However, we can simply integrate them with Monte Carlo. To illustrate this fact, we proceed as in the previous section.
The marginal $q(\Tf)$ is given by:
\begin{IEEEeqnarray}{rCl}
\begin{aligned}
        & q(\Tf) = \int q(\Tf,\Tu)d\Tu  \\
        & = \int p(\Tf|\Tu) q(\Tu) d\Tu   \\
        & = \int [\J{\Tf}-\J{\Tf|\Tu}\J{\Tu}^{-1}\J{\Tu|\Tf}]p(\Tfinit|\Tuinit) q(\Tuinit) d\Tuinit\\
        & =q(\Tfinit)\J{\Tf}- \int [\J{\Tf|\Tu}\J{\Tu}^{-1}\J{\Tu|\Tf}]p(\Tfinit|\Tuinit) q(\Tuinit) d\Tuinit
\end{aligned}\label{pepito}
\end{IEEEeqnarray}
where in the last line we use LOTUS rule. Note that integrating the inducing points will be generally intractable  due to the non-linearity of the flow $\G$ that appears in the conditional prior $p(\Tf|\Tu)$ through the elements $\J{\Tf|\Tu}$,$\J{\Tu|\Tf}$ and $\J{\Tu}$. Resorting to a Monte-Carlo approximation and simplifying the \ELL term in \ueqn\ref{apx_eqn:elbo}:
\begin{IEEEeqnarray}{rCl}
\begin{aligned}
    \ELL & = \sum_n\E_{q(\Tf,\Tu)} \left[ \log p(\Y_n \mid \Tfpos{\kk,n})\right] \\
    & = \sum_n\E_{q(\Tfpos{\kk,n},\Tu)}\left[ \log p(\Y_n \mid \Tfpos{\kk,n})\right] \\
    & \approx \sum_n  \frac{1}{S} \sum_s \left[ \log p(\Y_n \mid \Tfpos{\kk,n,s})\right]
\end{aligned}\nonumber
\end{IEEEeqnarray}
where we follow similar steps to marginal flows, i.e. we integrate out all the elements from $\Tf$ but $\Tfpos{\kk,n}$ (see below section). The last line is the Monte-Carlo approximation where samples are obtained by the generative process defined for flow based models, i.e \textit{sample from the base distribution} and \textit{warp samples with the flow}:
\begin{IEEEeqnarray}{rCl}
\begin{aligned}
   & \Tupos{0,s} \sim q(\Tuinit) \\
   & \Tfpos{0,n,s} \sim p(\Tfpos{0,n}|\Tupos{0,s})\\
   & \Tfpos{\kk,n,s},\Tupos{\kk,s} = \G_{\theta}(\Tfpos{0,n,s},\Tupos{0,s})
\end{aligned}
\end{IEEEeqnarray}
where the samples $\Tupos{\kk,s}$ are then discarded. 
Evaluation of this ELBO requires a computational complexity of $\BO(M^3)$ %
and although it requires two layers of sampling (from $q(\Tuinit)$ and $p(\Tfpos{0, n} \mid \Tuinit)$) this is similar to the doubly stochastic framework of \cite{salimbeni_dgp:2017} where samples must be propagated through the layers of the \DGP. Both sampling distributions are Gaussian and so the reparameterization trick can be used to generate low variance, unbiased gradients \citep{kingma_reparam_trick:2014}. Moreover, because $p(\Tfpos{0, n} \mid \Tuinit)$ is a univariate Gaussian sampling can be easily parallelized.
No matter the choice of $\G$ we have shown that we can always factorize the \ELL term across $\Y$ and $\Tf$ making this bound applicable to stochastic variational inference \citep{GPsBigData_hensman}. This allows the \ELL term to be approximated through mini-batching. Although this results in noisy gradient updates it allows $\TGP$ to be trained on millions of observations. 

\subsubsection{Computing the marginal distribution}

We now explicitly derive the marginal $q(\Tfpos{\kk, n})$ by integrating $\Tfpos{\kk,a\neq n}$ from $q(\Tfpos{\kk})$ from an alternative perspective. The derivation rests on a similar assumption made by \cite{VIinducingpoints_titsias} where $\Tu$ is assumed to be sufficient statistics for $\Tf$. Expanding the marginal $q(\Tfpos{\kk, n})$:
\begin{equation}
\begin{aligned}
	q(\Tfpos{\kk, n}) &= \int q(\Tfpos{\kk, n}, \Tfpos{\kk, a}) \text{d} \Tfpos{\kk, a} \\
	&=\int p(\Tfpos{\kk, n}, \Tfpos{\kk, a} \mid \Tu) q(\Tu) \text{d} \Tfpos{\kk, a} \text{d}\Tu \\
	& = \int p(\Tfpos{\kk, n} \mid \Tu, \Tfpos{\kk, a} ) p(\Tfpos{\kk, a} \mid \Tu) q(\Tu) \text{d} \Tfpos{\kk, a} \text{d}\Tu \\
   & =  \int p(\Tfpos{\kk, n} \mid \Tu ) q(\Tu) \text{d}\Tu  	
\end{aligned}
\end{equation}
For marginal flows this integration can be done analytically but for non-marginal flows we resort to Monte-Carlo approximations. The important point here is that no matter the form of $\G$ we just need the finite dimensional distributions at $\Z$ and $\X^{(n)}$ to evaluate the lower bound.

\subsection{Bayesian Input-Dependent Flows}
To finish with the mathematical derivations that complete the full specification of our variational posterior, we describe the necessary steps for marginal Bayesian input-dependent flows. The core idea is that the parameters of the flows that apply on each of the function evaluations $\Tfpos{0,n}$ at each index $\X^{(n)}$ of the stochastic process, depend directly on the index $\X^{(n)}$ -- rather than being shared across them.
Note that as the flow still applies over each of the function evaluations $\Tfpos{0,n}$ independently, then our input-dependent flows correspond to an input-dependent marginal flow. For arbitrary flows, one has to make sure that the necessary conditions are satisfied when the flow is made input-dependent. Hence this subsection just illustrates the derivation of the \ELBO within the derivation done for marginal flows. 

To do so, we assume independence between the stochastic process and the parameters of the Neural Network both on the prior and the posterior:
\begin{IEEEeqnarray}{rCl}
\begin{aligned}
   & p(\Tf,\Tu,\W) = p(\Tf|\Tu)p(\Tu)p(\W)\\
   & q(\Tf,\Tu,\W) = p(\Tf|\Tu)q(\Tu)q_{\phi}(\W)\\
\end{aligned}
\end{IEEEeqnarray}

where $\phi$ are the variational parameters of the \BNN. By plugging this into the \ELBO we arrive at:

\begin{IEEEeqnarray}{rCl}
\begin{aligned}
    \ELBO &= \E_{q(\Tf,\Tu)q(\W)}\log\left[\frac{\prod_n p(\Y_n|\Tfpos{\kk,n})p(\Tf,\Tu)p(\W)}{q(\Tf,\Tu)q_\phi(\W)}\right] \\
    &=\E_{q(\Tfinit)q_\phi(\W)}\log\left[\prod_n p(\Y_n|\Tfpos{\kk,n})\right] - \KL[q(\Tuinit)||p(\Tuinit)  - \KL[q_\phi(\W)||p(\W) \\
    & = \sum_n\E_{q(\Tfpos{0,n})q_\phi(\W)}\log\left[ p(\Y_n|\G_{\theta(\X,\W)}(\Tfpos{0,n}))\right]  - \KL[q(\Tuinit)||p(\Tuinit)] - \KL[q_\phi(\W)||p(\W)] \\
    &\approx \sum_n\frac{1}{S}\sum_s\E_{q(\Tfpos{0,n})}\log\left[ p(\Y_n|\G_{\theta(\X,\W_s)}(\Tfpos{0,n}))\right]  - \KL[q(\Tuinit)||p(\Tuinit)] - \KL[q_\phi(\W)||p(\W)] \\
\end{aligned}\nonumber
\end{IEEEeqnarray}

where  $W_s \sim q_\phi(\W)$. This bound has two interesting properties. First one can allow for low variance and unbiased gradients w.r.t $\phi$ by reparameterization (something satisfied for popular choices of $q(\W)$ such as the mean-field Gaussian family). Second, one can account for prior miss-specification by substituting the $\KLD$ for other divergences, which has been shown to improve the performance of this model \citep[see][]{GVI}. 

In our work however we have implemented the \BNN using Monte Carlo dropout \citep{MCdropout} as it can be more efficiently trained and also allow us to avoid some well-known problems of mean field VI such as variance under-estimation see e.g \cite{PMLRbishop}. Nevertheless, our bound can be efficiently trained regardless of the specification of the variational family by using batched matrix computations.

Finally, note that computing the forward passes through the Neural Network are independent of the computation of $q(\Tfinit)$. This means that one can parallelize the computation of $q(\Tfinit)$ and $\theta(\W_s,\X)$.
\subsection{Predictions}
\label{apx:predictions}
To make predictions we replace the true, unknown, posterior with the variational approximation. At a test location $\Xtest$, after integrating out the inducing points, we have:
\begin{IEEEeqnarray}{rCl}
\begin{aligned}
p(\Ytest|\Xtest,\X,\Y) &= \int  p(\Ytest|\Tftest)p(\Tftest|\X,\Y)d\Tftest \approx\\
& \approx \int  p(\Ytest|\Tftest)q(\Tftest)d\Tftest  \\
& \approx \int  p(\Ytest|\Tftest)q(\Tfinittest)d\Tfinittest  \\
\end{aligned}
\end{IEEEeqnarray}
where we have again make use of the LOTUS rule. This integral can be computed with quadrature. Note that for arbitrary flows, one can integrate the inducing points by the same sampling procedure we have already introduced in \usec\ref{sec:appendix_non_marginal_flows}. For the case of Bayesian input-dependent flows, we further approximate the integral using Monte Carlo:
\begin{IEEEeqnarray}{rCl}
\begin{aligned}
p(\Ytest|\Xtest,\X,\Y) &= \int  p(\Ytest|\Tftest,\theta(\W,\Xtest))p(\Tftest,\W|\X,\Y)d\Tftest \approx\\
& \approx \int  p(\Ytest|\Tftest)q(\Tftest)q(\W)d\Tftest d\W  \\
& \approx \frac{1}{S} \sum_s \int  p(\Ytest|\Tftest,\theta(\W_s,\Xtest))q(\Tfinittest)d\Tfinittest \\
\end{aligned}
\end{IEEEeqnarray}
where again each element of the Monte Carlo is a 1 dimensional quadrature integral. The whole process for both input and non-input-dependent flows can be easily parallelized with batched operations over matrices. Confidence intervals are obtained by sampling from the approximate posterior predictive and computing the relevant quantiles. Moment estimation and predictive test log likelihood is done by one dimensional quadrature, Monte Carlo estimation, and the logsumexp trick. We provide full derivations of the necessary estimators in our GitHub implementation.

\subsubsection{Predicting with \titleVWGP}
\label{sec:appendix_prediction_with_transformed_likelihod}

The confidence intervals of the predictive distribution with a transformed likelihood ($\T \neq \mathbf{I})$) are the transformed confidence intervals of $p(\T(\Y^*))$:
\begin{IEEEeqnarray}{rCl}
	\mathit{CI}(\Y^*) = \T^{-1}(\mathit{CI}(T(\Y^*)))
	\nonumber
\end{IEEEeqnarray}
Prediction with a transformed likelihood requires evaluating the inverse of $\T$ which in general requires approximation such as Newton-Raphson. 

\subsection{Alternative Variational Families}
\subsubsection{Derivation of \titleGSP}
\label{sec:appendix_gsp_derivation}

We use \GSP to denote the model which uses a Gaussian variational family instead of the transformed Gaussian proposed in the main paper. The joint model is the same as the \TGP:
\begin{equation}
    p(\MY, \Tf, \Tu) = p(\MY \mid \Tf) p(\Tf, \Tu)
\end{equation}
where $p(\Tf, \Tu)$ is given in \ueqn \ref{eqn:app_sparse_prior} as :
\begin{equation}
	p(\Tf, \Tu) = p(\Tfinit \mid \Tuinit) p(\Tuinit) \J{\Tf, \Tu}
\end{equation}
with $\J{\Tf, \Tu} = \priorjac{\Tf} \priorjac{\Tu}$. The derivation of the variational follows that of the \SVGP and \TGP. Let $q(\Tf, \Tu) = p(\Tf \mid \Tu) q(\Tu)$ be the Gaussian approximate posterior such that:
\begin{equation}
\begin{aligned}
    q(\Tu) &= \N(\Tu \mid \vm, \MS) \\
    p(\Tf \mid \Tu) &= \N(\Tf \mid \Kxz \iKzz \Tu, \Kxx - \Kxz \iKzz \Kzx)
\end{aligned}
\end{equation}
then the \ELBO is given by:
\begin{equation}
\begin{aligned}
\LL_{\GSP} &= \E_{q(\Tf, \Tu)} \left[ \log \frac{p(\MY \mid \Tf)p(\Tfinit \mid \Tuinit) p(\Tuinit) \J{\Tf, \Tu}}{p(\Tf \mid \Tu) q(\Tu)} \right]	\\
	&= \E_{q(\Tf)} \left[ \log p(\MY \mid \Tf) \right] + \E_{q(\Tf, \Tu)} \left[ \log \frac{p(\Tfinit \mid \Tuinit)}{p(\Tf \mid \Tu)} \right] + \E_{q(\Tf, \Tu)} \left[\log \frac{p(\Tuinit)}{q(\Tu)} \right] + \E_{q(\Tf, \Tu)} \left[ \log \J{\Tf, \Tu} \right] \\
	&= \E_{q(\Tf)} \left[ \log p(\MY \mid \Tf) \right] - \E_{q(\vu)} \big[ \KL \left[p(\Tf \mid \Tu) \mid \mid p(\Tfinit \mid \Tuinit) \right] \big] - \KL \left[ q(\Tu) \mid \mid p(\Tuinit) \right] + \E_{q(\Tf, \Tu)} \left[ \log \J{\Tf, \Tu} \right]
\end{aligned}
\end{equation}
The approximate posterior is no longer transformed by the same flow as the prior which means the Gaussian conditionals and Jacobian term no longer cancel. The first term is the same expected log likelihood term found in the \SVGP and so can be computed in closed form or with one dimensional quadrature as required. The second and fourth term must be approximated using monte-carlo due to the dimension of the base distribution.

Prediction using \GSP follows the standard \SVGP:
\begin{equation}
\begin{aligned}
	p(\MY^*) &= \intg{p(\MY^* \mid \Tf^*)p(\Tf^*)}{\Tf^*}{}{} \\
	&\approx \intg{p(\MY^* \mid \Tf^*)q(\Tf^*)}{\Tf^*}{}{}
\end{aligned}
\end{equation}
which can be computed using the \SVGP prediction equations.

\subsection{Other Computational Aspects}
To end this part of the appendix, we highlight and summarize some convenient computational aspects that derive from our approximation and what other modeling choices imply.

One of the most interesting properties of our model is that both training and predictions can be done without inverting $\G$ -- which allow us to use any expressive invertible transformation (see \uapendix \ref{apx:additional_results}). This contrasts with models that warp the likelihood, where either strong constraints must be placed on the kind of transformations employed so that the inverse can be computed analytically  \citep{compositionally_warped_gp_rios:2019}; or the inverse has to be computed using numerical methods as Newton-Raphson  \citep{snelson_warpedgp}.

On the other hand, other choices for the variational distribution would imply an undesirable increase in the computational time. First, the definition of $q(\Tu)$ allows us to compute the $\KL$ in closed form, avoiding the need to resort to estimation by sampling and to compute the Jacobian of the transformation. Note that computing this Jacobian can be done in linear time for marginal flows although it will have, in general, a cubic cost (see \cite{NFreview} for a review on the key points of Normalizing Flows).  Moreover, other choices of $q(\Tu)$ could require computation of the inverse $\G^{-1}_{\theta}(\Tu)$ to evaluate the density of the posterior sample under $p(\Tuinit)\J{\Tu}$. On the other hand, not canceling  $p(\Tf|\Tu)$ would also require to approximate the determinant of the transformation $\J{\Tf}$ and the inverse $\G_{\theta}$, with \textit{the whole dataset} $\X$, something that cannot be done stochastically. So canceling this term is not only important to allow \SVI, but also to avoid costly Jacobian/inverse computations. 

Finally, the use of marginal flows and definition of the variational posterior allow us to analytically integrate out the inducing points -- a very desirable property both for training and when making predictions.
\clearpage
\section{Additional Details on Experiments}
\label{apx:additional_results}

\subsection{Description of Flows}

In this section we provide a description of the flows used throughout the experiments. We describe the `base' transformations that are the building blocks to compositional and input-dependent flows, see Table. \ref{table:appendix_base_flow_definitions}.

\begin{table}[ht]
\caption{Description of Flows} \label{sample-table}
\begin{center}
\begin{tabular}{ccccc}
\textbf{Flow}  & \textbf{Forward} & \textbf{Inverse} & \textbf{Parameters}  \\
\hline \\
Log & $\log(\Tfinit)$ & $\exp(\Tf)$ &  ---   \\
Exp & $\exp(\Tfinit)$ & $\log(\Tf)$ &  ---   \\
Softplus         & $\log ( \exp (\Tfinit) + 1)$ & $\log ( \exp (\Tf) - 1)$ & ---  \\
Sinh & $\sinh(\Tfinit)$ & $\arcsinh(\Tfinit)$ & ---  \\
Arcsinh & $a \arcsinh(b(\Tfinit+c)) + d$ & $\sinh(\Tf)$ & $a, b,c,d \in \R$  \\
Affine (L) & $a + b \cdot \Tfinit$ & $\frac{\Tf-a}{b}$ & $a, b : \R$  \\ 
Sinh-Archsinh             & $\sinh(b \cdot \arcsinh(\Tfinit)-a)$ & $\sinh (\frac{1}{b}(\arcsinh(\Tfinit)+a))$ & $a, b : \R$  \\
Boxcox & $\frac{1}{\lambda} (\text{sgn}(\Tfinit) \mid \Tfinit \mid^{\lambda} -1)$ & $\text{sgn}(\lambda \cdot \Tfinit + 1) \mid \lambda \cdot \Tfinit + 1 \mid^{\frac{1}{\lambda}}$ & $\lambda > 0 $ \\
Tukey & $\frac{1}{g} \left[ \exp(g \cdot \Tfinit) -1 \right] \exp(\frac{h}{2} \Tfinit^2)$ & --- & $h \in R^+, g \in R : g \neq 0$   \\
TanH & $ a \tanh(b(\Tfinit + c)) + d $ & --- & $a, b,c,d \in \R$  \\
\end{tabular}
\end{center}
\label{table:appendix_base_flow_definitions}
\end{table}

These flows may combined to construct compositional flows and their parameters may be input-dependent. Moreover, we can create a new flow by linear combination of any of these flows, where the parameters have to be restricted so that each individual flows are strictly increasing/decreasing functions.  An example could be the following:
\begin{IEEEeqnarray}{rCl}
\Tf = \sum^I_i a_i + b_i\cdot\arcsinh((\Tfinit-c_i)/d_i);b_i,d_i \in \mathbb{R}^{+}\,\, \forall i
\end{IEEEeqnarray}
Finally, as in \cite{compositionally_warped_gp_rios:2019}  we define \SAL to be a Sinh-Archsinh flow with an Affine flow (L). We also consider compositions of flows made up directly by the inverse parametrization. For example in one of our experiments we experimented with the BoxCox+L flow and the InverseBoxCox+L flow. We provide this additional information in our GitHub.

\subsection{Initializing Flows from Data}

In this section we describe an initialization scheme that attempts to learn flow parameters that Gaussianize the prior. In the derivation we approximate the data as being noise-free and so in practice this may also be used for the likelihood transformations of the \WGP. Ideally we would want to learn a normalizing flow $\G(\cdot)$ that transforms a standard Gaussian $\varphi$ to the true prior $p(\vf_0)$:
\begin{equation}
	\varphi(\G^{-1}(\vf))\frac{\partial \G^{-1}}{\partial \vf} = p(\vf_0)
\label{eqn:gauss_eqn}
\end{equation}
In practice we do not have access to the true prior but instead observations $\Y$. By using $\Y$ as approximate samples from $p(\vf_0)$ we can then optimize $\G$ to approximately satisfy \ueqn \ref{eqn:gauss_eqn}. To optimize we directly minimize the KL divergence between $p(\vf_0)$ and $\varphi(\G^{-1}(\vf))\frac{\partial \G^{-1}}{\partial \vf}$: 
\begin{equation}
	\KL \left[ p(\vf_0) \mid \mid \varphi(\G^
	{-1}(\vf))\frac{\partial \G^{-1}}{\partial \vf} \right] = \E_{p(\vf_0)} \left[ \log \varphi(\G^
	{-1}(\vf))\frac{\partial \G^{-1}}{\partial \vf} \right] - \E_{p(\vf_0)} \left[ p(\vf_0) \right]
\end{equation}
The second term is constant w.r.t to the flow parameters and hence we only need to consider and optimize the first term. Because we have assumed that the $\MY$ are approximate samples from the true prior we write:
\begin{equation}
	E_{p(\vf_0)} \left[ \log \varphi(\G^
	{-1}(\vf))\frac{\partial \G^{-1}}{\partial \vf} \right] \approx \sum^N_{n=1} \log \varphi(\G^
	{-1}(\MY_n))\frac{\partial \G^{-1}}{\partial \MY_n} = \sum^N_{n=1} \log \varphi(\G^
	{-1}(\MY_n))(\frac{\partial \G}{\partial \G(\MY_n)})^{-1}
\end{equation}
and the final initialization optimization procedure is:
\begin{equation}
	\argmin_{\theta} \sum^N_{n=1} \log \varphi(\G^
	{-1}(\MY_n))\frac{\partial \G^{-1}}{\partial \MY_n}
\label{eqn:gauss_init_procedure}
\end{equation}
A  similar derivation is used by \cite{MAFflows} but in a different context.

\subsection{Initializing Flows approximately to Identity}

In this section we provide details on how we initialize flows close to identity. Many transformations can already recover identity but for those that cannot this method provides an effective and simple way to initialize them. To find these parameters we simply generate observations from the line $y=x: \left[ x_n, y_n\right]^N_{n=1}$ and minimize the MSE loss of the flow mapping from $x$ to $y$: 
\begin{equation}
	\argmin_\theta \frac{1}{N} \sum^N_{n=1} (y_n - \G(x_n))^2
\end{equation}

\subsection{ Initialization of Input-dependent flows}
\label{apx:subsec:inp_dep_initialization}
To initialize input-dependent flows we first initialize standard (non-input-dependent) flow parameters $\hat{\theta}$ by any of the procedures described above. Then, we turn the parameters into input-dependent and initialize the \NN parameters to match the values learned in the first step. This is done through stochastic gradient optimization, i.e. by first sampling a minibatch from the data distribution, and then minimize the empirical MSE loss between $\NN(\X)$ and $\hat{\theta}$.

\subsection{Real World Experiments}

For both real world experiments, we consider 2 different seeds, shuffle the observations and run 5-fold cross-validation across 2 different optimization schemes. The first optimization scheme optimizes both the variational and hyperparameters jointly. The second holds the likelihood noise fixed for 60\% of iterations. This is to help avoid early local minimum that causes the models to underfit and explain the observations as noise.

For all models, we use RBF kernels with lengthscales initialized to 0.1, and Gaussian likelihoods with noise initialized to 1.0. We optimize the whitened variational objective using Adam optimizer with a learning rate of 0.01.

\subsubsection{Air Quality}

We used data from the London Air Quality Network \cite{laqn} and we focus on site HP5 (Honor Oak Park , London) using 1 month of PM25 data (731 observations, date range 03/15/2019 - 04/15/2019). Because the observations are non-negative bounded we only consider the following positive enforcing flows:  sal+softplus randomly initialized, sal+softplus initialized from data and a sal + sal + softplus initialized from data.

We shuffle observations and run 5-fold cross validation across 5\%, 10\% and 100\% of inducing points and optimize each for a total of 10000 epochs. We compute means and standard deviations across all flows, folds and seeds. In the main paper we only present results using a sal+softplus (initialized from data) flow and we now present additional results across all these positive enforcing flows.

\textbf{Additional results averaged across multiple flows}

We now additionally present the results averaged across all the considered flows. The results echo the findings on the single flow experiment, that the \TGP outperforms both the \GP and \VWGP across the 3 levels of inducing points considered. This suggests that the results of the \TGP are somewhat invariant to the choice of (positive enforcing) flows used.

\begin{table}[ht]
	\begin{minipage}{.5\linewidth}	
		\centering
		\captionof{figure}{RMSE (left is better) on the Air quality dataset across 5\%, 10\% and 100\% of inducing points.  }
		\includegraphics{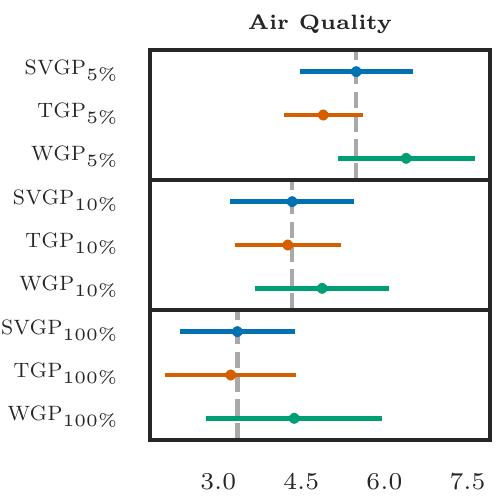}
	\end{minipage}%
	\begin{minipage}{.5\linewidth}
	    \captionof{table}{Table of results to reproduce the left panel}
		\begin{tabular}{ccc}
			\toprule
			 Model   &   RMSE &   Standard Deviation \\
			\midrule
			 GP 5    &  5.491 &                1.02  \\
			 TGP 5   &  4.896 &                0.713 \\
			 WGP 5   &  6.393 &                1.235 \\
			 GP 10   &  4.335 &                1.115 \\
			 TGP 10  &  4.257 &                0.961 \\
			 WGP 10  &  4.876 &                1.214 \\
			 GP 100  &  3.345 &                1.036 \\
			 TGP 100 &  3.227 &                1.18  \\
			 WGP 100 &  4.372 &                1.589 \\
			\bottomrule
		\end{tabular}
	\end{minipage}
\label{table:appendix_aq_additional_results_2}
\end{table}

\subsubsection{Rainfall}

The Switzerland rainfall uses data collected on the 8th of May 1986. Because all the observations are positively bounded we again use positive enforcing flows. We consider: softplus, sal+softplus (from data), sal+softplus (from identity), sal+sal+softplus (from identity), sal+sal+softplus (from data). 

We optimize for a total of 20000 epochs and compute means and standard deviations across all flows, folds and seeds. 

\begin{table}[ht]
	\begin{minipage}{.5\linewidth}	
		\centering
		\captionof{figure}{RMSE (left is better) on the Rainfall experiment.}
		\includegraphics{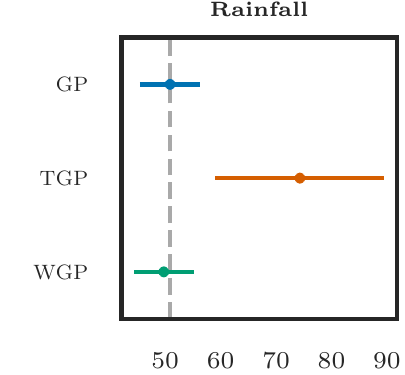}
	\end{minipage}%
	\begin{minipage}{.5\linewidth}
	    \captionof{table}{Table of results to reproduce the left panel.}
		\begin{tabular}{ccc}
			\toprule
			 Model   &   RMSE &   Standard Deviation \\
			\midrule
 GP      & 50.892 &                5.441 \\
 TGP     & 74.306 &               15.225 \\
 WGP     & 49.764 &                5.369 \\
			\bottomrule
		\end{tabular}
	\end{minipage}
\label{table:appendix_aq_additional_results}
\end{table}

\subsubsection{Table of results in figures in main paper}

In this section we present, for reproducibility and comparison reasons, the numerical values used to generate \fig6 in the main paper.

\begin{table}[ht]
\caption{Table of results to generate \fig 6 in Main paper. The left table shows results for the Air quality experiment (Left panel in \fig 6) and the right table shows results for the rainfall experiment (Right panel in \fig 6).}
\begin{center}
	\begin{minipage}{.5\linewidth}
		\centering
		\begin{tabular}{lrr}
		\toprule
		 Model   &   RMSE &   Standard Deviation \\
		\midrule
		 GP 5    &  5.491 &                1.02  \\
		 TGP 5   &  4.851 &                0.786 \\
		 WGP 5   &  6.451 &                1.207 \\
		 GP 10   &  4.335 &                1.115 \\
		 TGP 10  &  4.299 &                1.269 \\
		 WGP 10  &  4.505 &                0.986 \\
		 GP 100  &  3.345 &                1.036 \\
		 TGP 100 &  2.833 &                0.923 \\
		 WGP 100 &  4.151 &                1.702 \\
		\bottomrule
		\end{tabular}
	\end{minipage}%
	\begin{minipage}{.5\linewidth}
		\centering
		\begin{tabular}{lrr}
		\toprule
		 Model      &   RMSE &   Standard Deviation \\
		\midrule
		 GP         & 50.892 &                5.441 \\
		 TGP SP     & 50.858 &                5.395 \\
		 WGP SP     & 50.893 &                5.356 \\
		 TGP SAL+SP & 73.057 &                8.465 \\
		 WGP SAL+SP & 48.85  &                5.404 \\
		\bottomrule
		\end{tabular}
	\end{minipage}%
\end{center}
\end{table}

\subsection{Black Box Results}

In the black box experiments we explore the performance of \TGP across many UCI datasets \citep{UCI}. The performance measures are evaluated by employing random 10 fold train-test partitions and reporting average results plus standard error. This is done for all the datasets except Year and Avila (because the test partition is already provided) and Airline, where we just use a random 1 fold partition following previous works e.g \citep{salimbeni_dgp:2017}.
We perform flow selection by running each of the candidate models using random validation splits. We use one validation split on the first and second random fold partitions, except for Year and Airline where we only use one. This selection is done for 100 inducing points. The selected model is then used across all the experiments reported, including the experiments with less inducing points.

To initialize our models we follow \citet{salimbeni_dgp:2017}. We use RBF kernels with parameters initialized to $2.0$. The inducing points are initialized using the best of 10 KMeans runs except for Year and Airline where we just use 1 run. We use a whitened representation of inducing points and initialize the variational parameters to $\vm=0$ and $\MS = 1^{-5} I$. The \DGP have an additional white noise kernel added to the RBF in each hidden layer, with the noise parameter initialized to  $1^{-6}$. The noise parameter of the Gaussian likelihood is initialized to $0.05$ for the regression experiments. For classification we use a noise free latent function and Bernoulli/Categorical likelihoods for Binary/Multiclass problems. We use probit and softmax link functions respectively. We use Adam optimizer with a learning rate of $0.01$. The flow initializers are run over 2000 epochs for the identity initializer and over 2000 (rest-of-datasets)  or 20 (Year-Airline) epochs for input-dependent flows, to match the learned parameters in the previous initialization step. In our experiments however, we observed that the flow could be initialized with fewer epochs. We have tried a set of different combinations of flows as described in \utable \ref{table:appendix_base_flow_definitions}. This includes different flow lengths and different number of flows in the linear combinations. For input-dependent flows we just tried the SAL flow with lengths 1 and 3; where dependency is encoded just in the non-linear flow (i.e the sinh-arcsinh). For these experiments we focus on exploring different architectures of the Neural Networks. We search over $\{1,2\}$ hidden layers, $\{25,50\}$ neurons per layer, $\{0.25,0.5,0.75\}$ dropout probabilities, batch normalization  and ReLu and Tanh activations.  We found that any of these possible combinations could work except the use of Batch Normalization, and that most of this combinations could be successfully optimized using the default optimizer, although some combinations suggested that a lower learning rate was needed to make optimization stable (this combinations were discarded as we wanted to show robustness against optimizer hyperparameter search). The prior over the neural network parameters is kept fixed and is introduced in the model by fixing a $1^{-5}$ weight decay in the optimizer ($\lambda=1^{-5}$). In our Github we provide additional information about the model selection process and the final selected models.  
All of our models were optimized for 15000 epochs for all datasets except Year and Airline where we use 200 epochs. Each epoch corresponds to a full pass over the dataset. For classification we follow \cite{classificationsparseGPVIHensman} and freeze the covariance parameters before learning everything end to end. This is done for the first 2000 epochs. 

For experiments using less than 100 inducing points we use the same flow architecture selected from the validation sets and optimizer hyperparameters as the corresponding 100 inducing point experiment.
The performance obtained highlights that our approach is somewhat robust to hyperparameter selection. On just one dataset we observe that this extrapolation was suboptimal and that the algorithm diverge. Those results are not reported in this appendix and correspond to 5 inducing points, input-dependent flows and \texttt{naval} dataset. We attribute this fact to not having performed model selection and optimizer hyperparameter search for each set of inducing points specifically.
On the other hand, if in any of the experiments carried out failed due to numerical errors (e.g Cholesky decomposition) we increase the standard amount of jitter added by Gpytorch from $1^{-8}$  to $1^{-6}$. This is just needed on some train-test folds and some datasets only when using less inducing points. In general we found that our experiments were quite stable. 

\subsubsection{Regression}
In this subsection we present the complete results across the regression benchmarks. This includes the \NNL and \RMSE for different numbers of inducing points on the following models: Sparse Variational \GP (\SVGP), Transformed \GP with non input-dependent flows (\TGP), Transformed \GP with point estimate input-dependent flows (\PETGP) and Transformed \GP with Bayesian input-dependent flows (\BATGP). Note that \PETGP and \BATGP share the same input-dependent architecture and learned parameters $\W$, and the only difference relies on whether we use standard Dropout \citep{standardDropout} or Monte Carlo dropout \citep{MCdropout} to make predictions.  

\paragraph{Large Scale Regression} We report the \NLL and \RMSE for the Large scale regression problem in \fig \ref{fig:ALL_LL_large_regression} and \fig \ref{fig:ALL_RM_large_regression}. Across both datasets our model outperforms the baseline \GP both in \RMSE and \NNL. Furthermore, the \TGP also performs well, although in general is outperformed by the \BATGP. These plots also show the regularization effect of Bayesian marginalization. In Year the \RMSE and \NNL is highly improved when accounting for uncertainty in the parameters. In Airline both models provide similar \RMSE, but the \BATGP provides better uncertainty quantification, reflected by improved \NNL scores. In these experiments the \DGP mostly outperforms \TGP,  \PETGP and \BATGP except for on the Year dataset where \BATGP has a lower \NNL, indicating better uncertainty quantification.

\begin{figure}[H]
\begin{subfigure}[t]{\textwidth}
      \centering
    \includegraphics[width=0.5\textwidth]{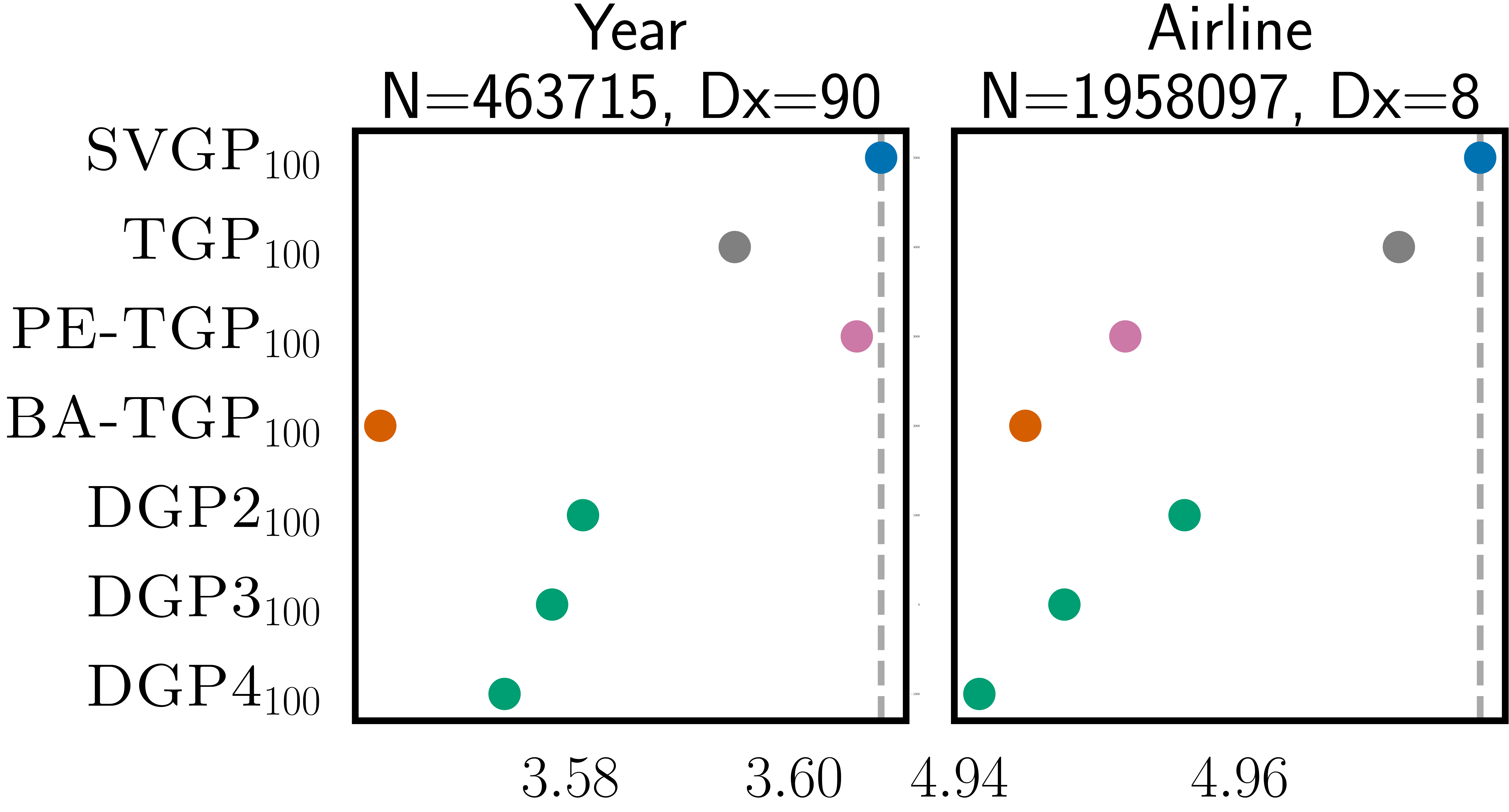}
    \caption{\NNL (left is better) for large scale regression datasets.}
    \label{fig:ALL_LL_large_regression}
\end{subfigure}\vspace{0.2cm}
\begin{subfigure}[t]{\textwidth}
      \centering
    \includegraphics[width=0.5\textwidth]{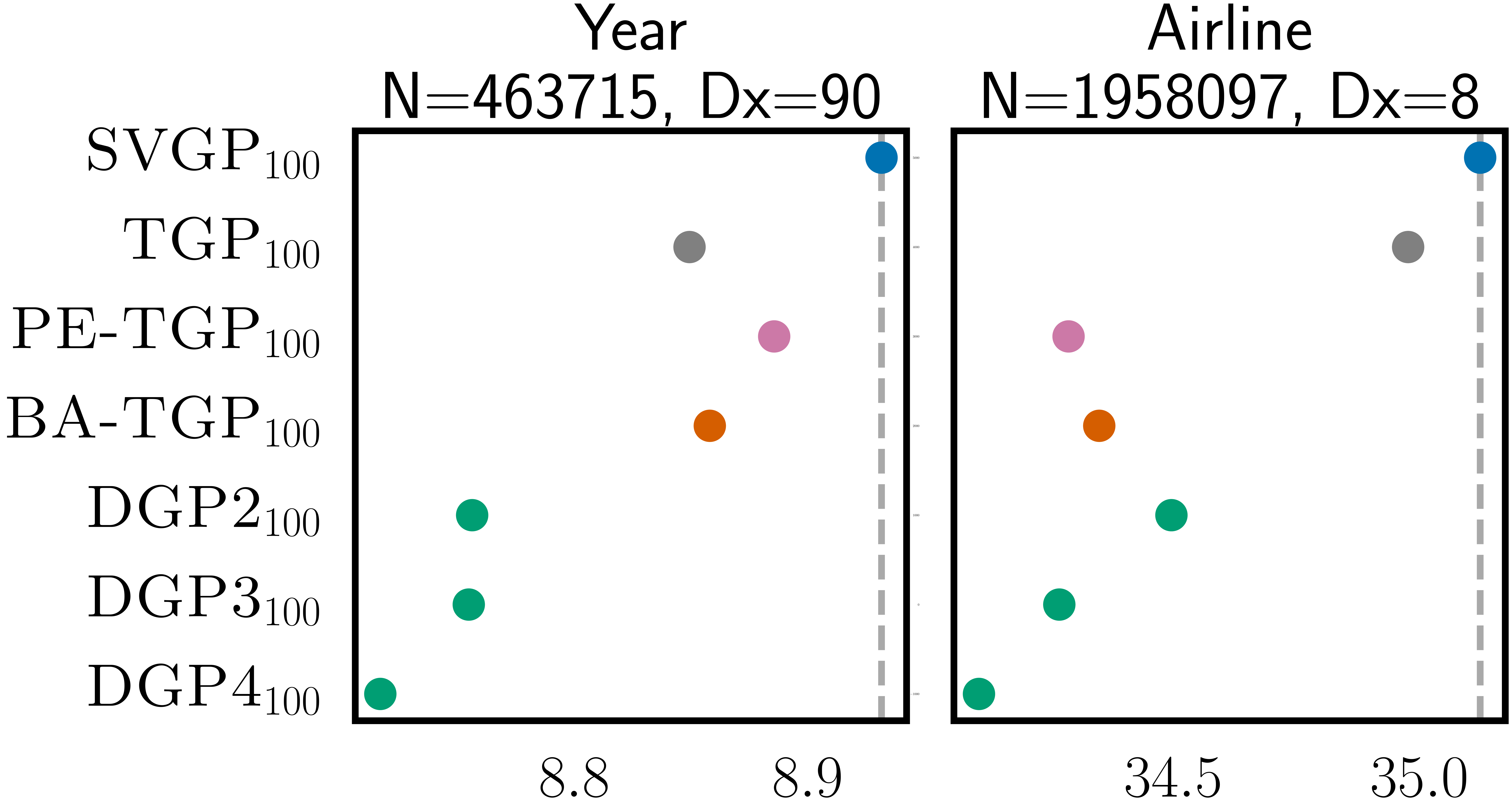}
    \caption{\RMSE (left is better) for large scale regression datasets.}
    \label{fig:ALL_RM_large_regression}
\end{subfigure}
\begin{subfigure}[t]{\textwidth}
      \centering
    \includegraphics[width=0.5\textwidth]{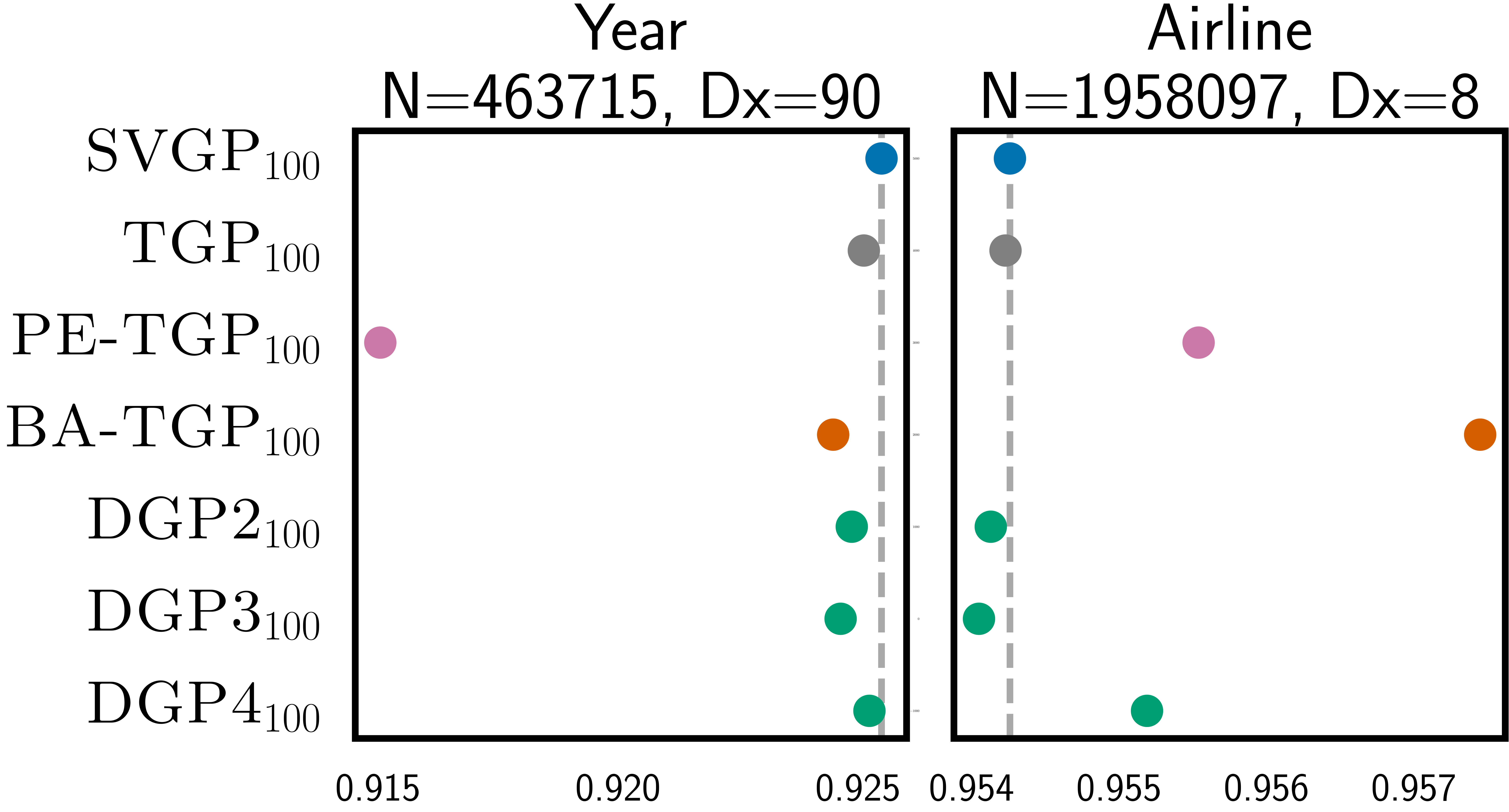}
    \caption{ \COV (right is better) for large scale regression datasets.}
    \label{fig:ALL_COV_large_regression}
\end{subfigure}
\end{figure}

\paragraph{Medium-Small Regression} We now illustrate the results for the medium small regression in \fig \ref{fig:ALL_LL_medium_small_regression}  (\NNL) and \fig \ref{fig:ALL_RMSE_medium_small_regression} (\RMSE). We show results split across decreasing number of inducing points and different models.  Across all levels of inducing points, the \BATGP model ranks the best and consistently outperforms alternative models on both \NNL and \RMSE showing superior point-prediction and uncertainty quantification. 

On the other hand, by looking at e.g \texttt{power} dataset, and comparing \RMSE and \NNL, we can see how in terms of \RMSE both the \PETGP and \BATGP perform similarly. However, there is a big difference in terms of \NNL, which is an indicator of good predictive uncertainty quantification provided by introducing uncertainty in the flow parameters. We build on this observation in the final subsection of this appendix. 

Moreover, a particularly interesting outcome is the performance of the models when only using 5 inducing points. We can see that in \texttt{kin8nm}, \texttt{power} and \texttt{concrete} the 5 inducing points provides a similar performance to the 100 inducing points for the \BATGP. We show in the subsequent section that even though the Neural network is highly expressive, the base \GP is necessary and not `ignored' by the model. Hence we can attribute the excellent performance of \BATGP to both the combination of the \BNN and the \GP.

We can also see how the standard \TGP is also able to improve upon the \GP in some datasets, although the improvement is minimal, clearly highlighting the necessity of input-dependent flows. The fact that the standard \TGP has been tested using more complex transformations than the input-dependent \TGP (which uses just 1-3 length \SAL flows) suggest that the boost in performance clearly comes from the input dependency and not the warping function. This means that these results can be improved by testing more complex input-dependent flow combinations, which is something we let for future work. Also we can see in \texttt{wine-red} that while the uncertainty quantification of our model and the \DGP is quite similar, we clearly outperform it in terms of pointwise predictions. The \DGP consistently outperforms the \TGP and \SVGP  w.r.t \RMSE but in general the \BATGP and \PETGP achieve superior performance. These two observations might indicate that the proposed model is more expressive in terms of pointwise predictions than a \DGP.

Finally, note how without doing specific model selection for the less inducing points models, the parameters extrapolated from the 100 inducing points one works very well, which means that our model is somewhat robust to the selection of hyperparameter. Also by noting that all the models are trained for 15000 epochs, we show how our model has not over-fit, although being much more complex than a `simple' \GP.

\begin{figure}[!t]
\begin{subfigure}[t]{\textwidth}
      \centering
       \includegraphics[height=0.42\textheight]{images/appendix_B/ALL_RESULTS_MEDIUM_SMALL_REGRESSION/LL_boston_energy_concrete_wine-red_wine-white_models_GP_TGP_PE_TGP_BA_TGP_DGP_2_DGP_3_DGP_4_num_Z_100_50_20_10_5.pdf}
\end{subfigure}\vspace{0.0cm}
\begin{subfigure}[t]{\textwidth}
      \centering
        \includegraphics[height=0.42\textheight]{images/appendix_B/ALL_RESULTS_MEDIUM_SMALL_REGRESSION/LL_kin8nm_power_naval_protein_ranking_models_GP_TGP_PE_TGP_BA_TGP_DGP_2_DGP_3_DGP_4_num_Z_100_50_20_10_5.pdf}
    \end{subfigure}
     \caption{ Comparing \NNL (left is better) across 9 data sets for several number of inducing points. \textbf{Bottom right panel:} Ranking of the methods across all 9 data sets. \TGP stands for non input-dependent flows, \PETGP stands for point estimate input-dependent flows (Standard Dropout) and \BATGP stands for the Bayesian input-dependent flows (MC Dropout) }
     \label{fig:ALL_LL_medium_small_regression}
\end{figure}

\begin{figure}[!t]
\begin{subfigure}[t]{\textwidth}
      \centering
       \includegraphics[height=0.42\textheight]{images/appendix_B/ALL_RESULTS_MEDIUM_SMALL_REGRESSION/RM_boston_energy_concrete_wine-red_wine-white_models_GP_TGP_PE_TGP_BA_TGP_DGP_2_DGP_3_DGP_4_num_Z_100_50_20_10_5.pdf}
\end{subfigure}\vspace{0.0cm}
\begin{subfigure}[t]{\textwidth}
      \centering
        \includegraphics[height=0.42\textheight]{images/appendix_B/ALL_RESULTS_MEDIUM_SMALL_REGRESSION/RM_kin8nm_power_naval_protein_ranking_models_GP_TGP_PE_TGP_BA_TGP_DGP_2_DGP_3_DGP_4_num_Z_100_50_20_10_5.pdf}
    \end{subfigure}
     \caption{ Comparing \RMSE (left is better) across 9 data sets for several number of inducing points. \textbf{Bottom right panel:} Ranking of the methods across all 9 data sets. \TGP stands for non input-dependent flows, \PETGP stands for point estimate input-dependent flows (Standard Dropout) and \BATGP stands for the Bayesian input-dependent flows (MC Dropout).}
     \label{fig:ALL_RMSE_medium_small_regression}
\end{figure}

\begin{figure}[!t]
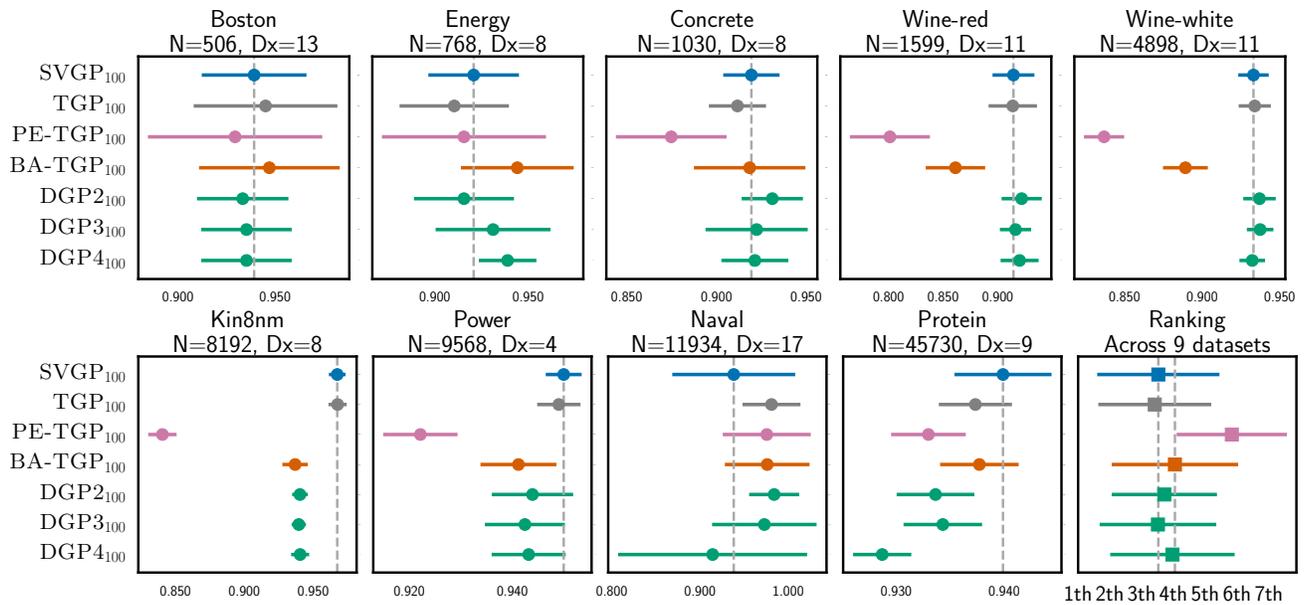

\begin{subfigure}[t]{\textwidth}
      \centering
       \includegraphics[width=\textwidth]{images/appendix_B/ALL_RESULTS_MEDIUM_SMALL_REGRESSION/CO_boston_energy_concrete_wine-red_wine-white_models_GP_TGP_PE_TGP_BA_TGP_DGP_2_DGP_3_DGP_4_num_Z_100.pdf}
\end{subfigure}\vspace{0.0cm}
\begin{subfigure}[t]{\textwidth}
      \centering
        \includegraphics[width=\textwidth]{images/appendix_B/ALL_RESULTS_MEDIUM_SMALL_REGRESSION/CO_kin8nm_power_naval_protein_ranking_models_GP_TGP_PE_TGP_BA_TGP_DGP_2_DGP_3_DGP_4_num_Z_100.pdf}
    \end{subfigure}
     \caption{Comparing \COV (right is better) across 9 data sets for 100 inducing points. \textbf{Bottom right panel:} Ranking of the methods across all 9 data sets. \TGP stands for non input-dependent flows, \PETGP stands for point estimate input-dependent flows (Standard Dropout) and \BATGP stands for the Bayesian input-dependent flows (MC Dropout).}
     \label{fig:100_COV_medium_small_regression}
\end{figure}

\clearpage

\subsubsection{Classification}

In this section we expand on the classification experiments from the main paper by providing two additional datasets and reporting accuracy. For all the datasets we show \NLL in \fig \ref{fig:LL_classification} and  accuracy in \fig \ref{fig:acc_classification}. Across all datasets our proposed models (either through input-dependent or non-input-dependent flows) outperform the \SVGP, and only in \texttt{heart} does the \TGP substantially outperform the input-dependent models \BATGP and \PETGP. We highlight the big boost in accuracy provided by our models (see e.g \texttt{avila} dataset).
Surprisingly we observe that the \PETGP performs well in classification and usually outperforms the \BATGP. However, we have observed that sometimes the \PETGP outputs extreme wrong values, giving a \NNL of $\infty$. For this same model we observe that the \BATGP was able to remove those extreme predictions. We speculate that the model is correctly incorporating epistemic uncertainty relaxing extremely wrong assigned confidences. 
On the other hand, we attribute the bad performance of the \BATGP and \PETGP in the \texttt{heart} dataset to prior misspecification. First note that L in this case the\NN is not depicted for the \PETGP, as this is one of the cases in which many predictions were extremely wrong, yielding a $\infty$ \NNL. We can see how the \BATGP solves this, but due to this misspecification is unable to provide good predictions. As explained by \cite{GVI}, prior misspecification leads to a misleading quantification of uncertainty. 
Further, the number of training data points is small, meaning that the prior has a relatively strong influence relative to the likelihood terms. 
In situations like this, a badly specified prior dominates the likelihood terms and adversely affects the predictive likelihoods. 

To build on this claim we note that the \TGP outperforms the \GP on this dataset indicating that having a more expressive prior is beneficial. This coupled with the fact that we did not tune prior $p(\W)$ for our \BNN, and that this is the only dataset in which the \BATGP does not give a clear boost in performance, are consistent observations with the hypothesis of prior misspecification.

Finally on \texttt{banknote}, we see that all the models provide a 100\% accuracy, which means that the improvement in the \NNL is coming from reducing the calibration gap, as the \NNL is a proper scoring rule. Our model is making the predictions extreme towards the correct class, which is a desirable property if your data distribution doesn't present overlap between classes. Note however how in this case the \BATGP still provides uncertainty in the predictions, avoiding the extreme $\{0,1\}$ probability assignments.
 
\begin{figure}[H]
\begin{subfigure}[t]{\textwidth}
      \centering
    \includegraphics[width=\textwidth]{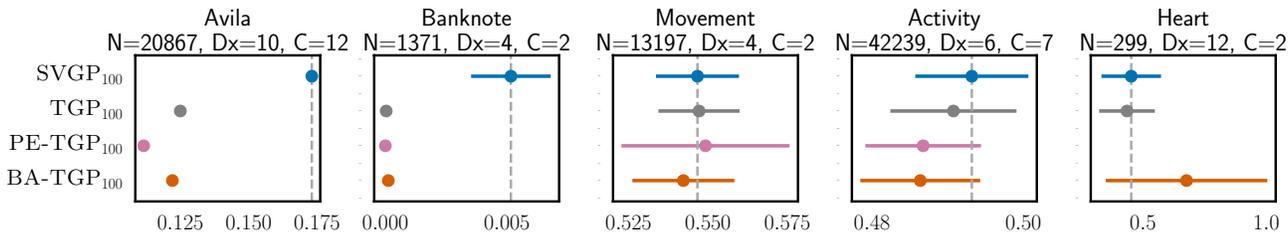}
    \caption{\NNL (left is better) of all the classification datasets}
    \label{fig:LL_classification}
\end{subfigure}\vspace{0.2cm}
\begin{subfigure}[t]{\textwidth}
      \centering
    \includegraphics[width=\textwidth]{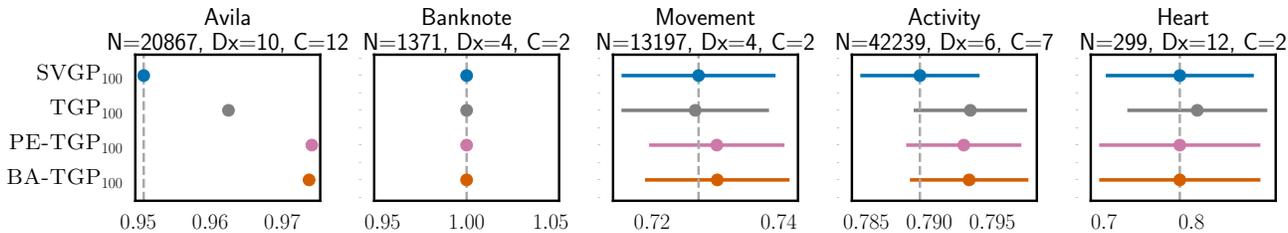}
    \caption{ Accuracy (right is better) of all the classification datasets}
    \label{fig:acc_classification}
\end{subfigure}
\caption{Results for classification datasets. \TGP stands for non input-dependent flows, \PETGP stands for point estimate input-dependent flows (Standard Dropout) and \BATGP stands for the Bayesian input-dependent flows (MC Dropout).}
\end{figure}

\subsubsection{Inference with Variational Bayes}
\label{apx:variational_bayes}
We run a subset of the experiments using variational bayes inference instead of Monte Carlo dropout, using a standard normal prior $p(\W)=\prod^{|\W|}_{\textbf{w}_i}\mathcal{N}(\textbf{w}_i|0,1)$ over the weights of the Neural Network and the same Neural Network architecture used by the MC dropout experiments to parameterize the input dependency (with the exception of the dropout layer that is removed).

We initialize the flows following the proceedure described in \usec{} \ref{apx:subsec:inp_dep_initialization}: The variational parameters from $q(\W)$ are optimized to minimize the sum of squared errors between the output of the neural network and the value of the parameter that makes the flow learn an identity mapping, i.e. we do not incorporate the $\KLD$ penalty from the \ELBO at this point. We found that this initialization procedure is harder to optimize as opposed to Monte Carlo dropout, and we let this for future work,  as this could depend not only on the inference method but on the kind of flow we use (1-3 length \SAL flows).

As a result of the initialization most of the models did not achieve a good starting point for the subsequent optimization and this makes the model either provide suboptimal results or directly make optimization unstable. We run experiments on the following datasets using just 5 seeds: \texttt{boston}, \texttt{energy}, \texttt{concrete}, \texttt{kin8nm} and \texttt{wine\_red}. We found that in all of them but \texttt{boston} and \texttt{concrete} the model saturated, and that in \texttt{concrete} the optimization was very unstable (even if we lower the learning rate), which leads to bad local optima. As shown in table below this made the model provide  unseful predictions (around $16.x$ values of \RMSE).

Beyond the initialization procedure the reasons behind these problems can be attributed to only using one Monte Carlo sample to estimate the \ELBO during training. 
On the other hand, using more Monte Carlo samples for training implies higher training time, while we are interested in efficient implementations of our model that are much faster than a \DGP. Furthermore, the results showed in the next table suggest that beyond this computational and optimization issues, MC dropout seems to provide better results than Variational Bayes under the same training specifications (1 sample for training, same likelihood parameterization etc), hence the study of possible alternatives for inference is something we leave for future work.

\begin{table}[H]
\centering
\begin{tabular}{cccccc}
                          &      & \SVGP & Point Estimate \TGP & MC dropout \TGP & Variational Bayes \TGP \\\hline
\multirow{2}{*}{Concrete} & \NLL  & 3.17 & 3.24           & 3.02       & 5.67              \\
                          & \RMSE & 5.79 & 5.55           & 5.57       & 18.02             \\\hline\hline
\multirow{2}{*}{Boston}   & \NLL  & 2.38 & 2.26           & 2.24       & 2.39              \\
                          & \RMSE & 2.66 & 2.38           & 2.33       & 2.60\\\hline             
\end{tabular}
\end{table}

\subsection{Bayesian Flows}

In this subsection  we provide additional information about Bayesian flows to highlight and connect with the conclusions provided in the UCI datasets section. We first show the effect of considering parameter uncertainty in the flow. We then illustrate why the whole modeling is not done by the \BNN.

\subsubsection{Uncertainty in the Warping Function}

To illustrate the parameter uncertainty introduced by using a \BNN we plot the warping function evaluated at four different training locations $\X^{(n)}$ for the \texttt{power} dataset. This means that we will show four different warping functions $\G$ with parameters given by $\{\theta(\W,\X^{(i)})\}^4_{i=1}$. Note that the weights of the $\W$ are shared and the difference in each of the parameters comes from the specifics inputs $\X$. The figures show the function parameterized by the different warping functions when applied on $\Tfinit$, i.e. we plot $\Tf=\G_{\theta(\W,\X^{(n))}}(\Tfinit)$. The range  $\{\Tfpos{0_a},\Tfpos{0_b}\}$  in which the flow is evaluated goes from the minimum $\Y$ value in the training dataset to the maximum one. In this way we show what kind of warping function has the model learned for the output range to be regressed. 
\begin{figure}[!b]
    \centering
    \includegraphics[width=\textwidth]{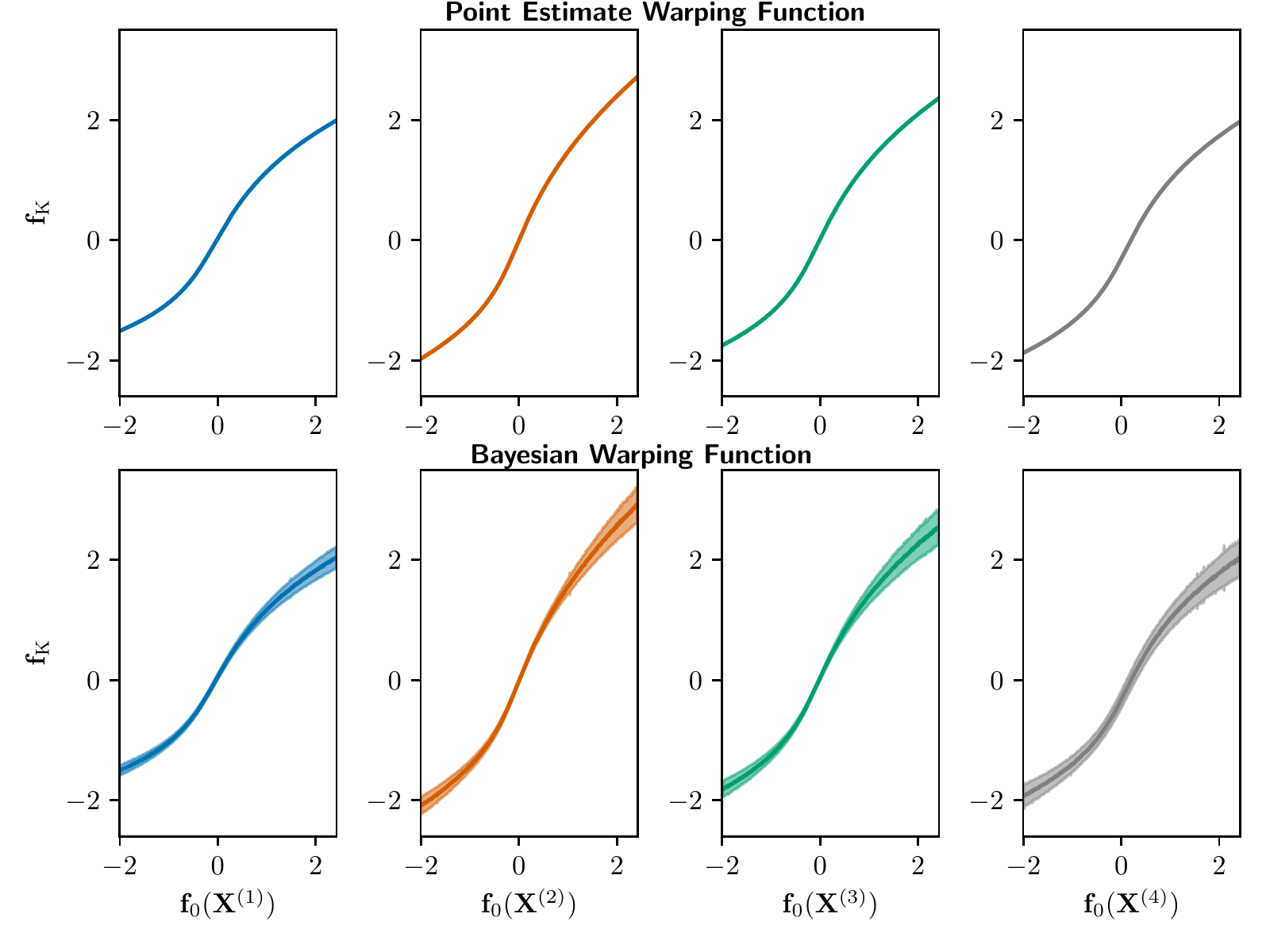}
    \caption{Example of warping functions obtained with input-dependent flows for the \texttt{power} dataset. The top row shows the point estimate warping function evaluated over a range $\Tfinit$, at different input locations using standard Dropout. The bottom row shows mean and standard deviation of samples from the posterior of the Bayesian flow using Monte Carlo Dropout. We can see how the model learns a different function warping depending on the input locations and how the model accounts for parameter uncertainty.}
    \label{fig:bayesian_flow}
\end{figure}

The top row in \fig \ref{fig:bayesian_flow} shows the transformation $\G_{\theta(\X^{(n)},\W)}$ for the point estimate flow. The bottom row shows the mean and the standard deviation of the same flow where the parameters have been sampled from the posterior distribution. Note that the top and bottom rows use the same Neural Network parameters. The only difference is that while in the top row we compute the flow parameters with one forward pass through the Neural Network, by multiplying the activations by the 1 minus the probability of dropout $p$ \citep{standardDropout}, in the bottom row we drop activations with probability $p$ on each forward through the Neural Network \citep{MCdropout}.

First, we can see how the model learns a different warping function for each different input $\X^{(n)}$, and this means that the marginal distribution at each index of our stochastic process is different. Second, we observe how the Bayesian flow incorporates parameter uncertainty. Note that both the Bayesian and point estimate flows have the same mean function; and this is the reason why the \RMSE metrics reported in \fig \ref{fig:ALL_RMSE_medium_small_regression} for the \texttt{power} dataset are very similar  (note that \RMSE only considers the first moment of the posterior predictive). However the \NNL showed in \fig \ref{fig:ALL_LL_medium_small_regression} is much better for the Bayesian flow, as we are incorporating epistemic uncertainty, hence being less confident in regions of function evaluations $\Tfinit$ where the model is uncertain. Third, note that the functions plotted for each of the training samples are simple, because the functional form is just given by a $K=3$ input-dependent flow. Hence input-dependent flows are not overfitting because the functional transformation $\G$ is complex, but because it is very specific for a particular point $\X^{(n)}$. By incorporating parameter uncertainty in the flow parameters we are considering all the possible functions parameterized by the flow's functional $\G$, hence preventing overfitting. We can expect this regularization effect to be bigger when making the functional's form more complex, which is something we let for future work.

\subsubsection{Uncertainty handled by the \titleGP{}}

In the UCI experiment section we comment that sometimes the model with 5 inducing points provides similar results to the 100 inducing points models. Initially we could think that the \BNN is handling all the modeling power both in terms of uncertainty quantification and regressed values. In this subsection we illustrate that this is not the case and that the modeling performance comes from a combination of the \NN and the \GP{}. Note that the uncertainty provided by the \GP{} is combined with the uncertainty provided by the \BNN.

\begin{figure}[!t]
\begin{subfigure}[t]{0.5\textwidth}
    \centering
    \includegraphics[width=\textwidth]{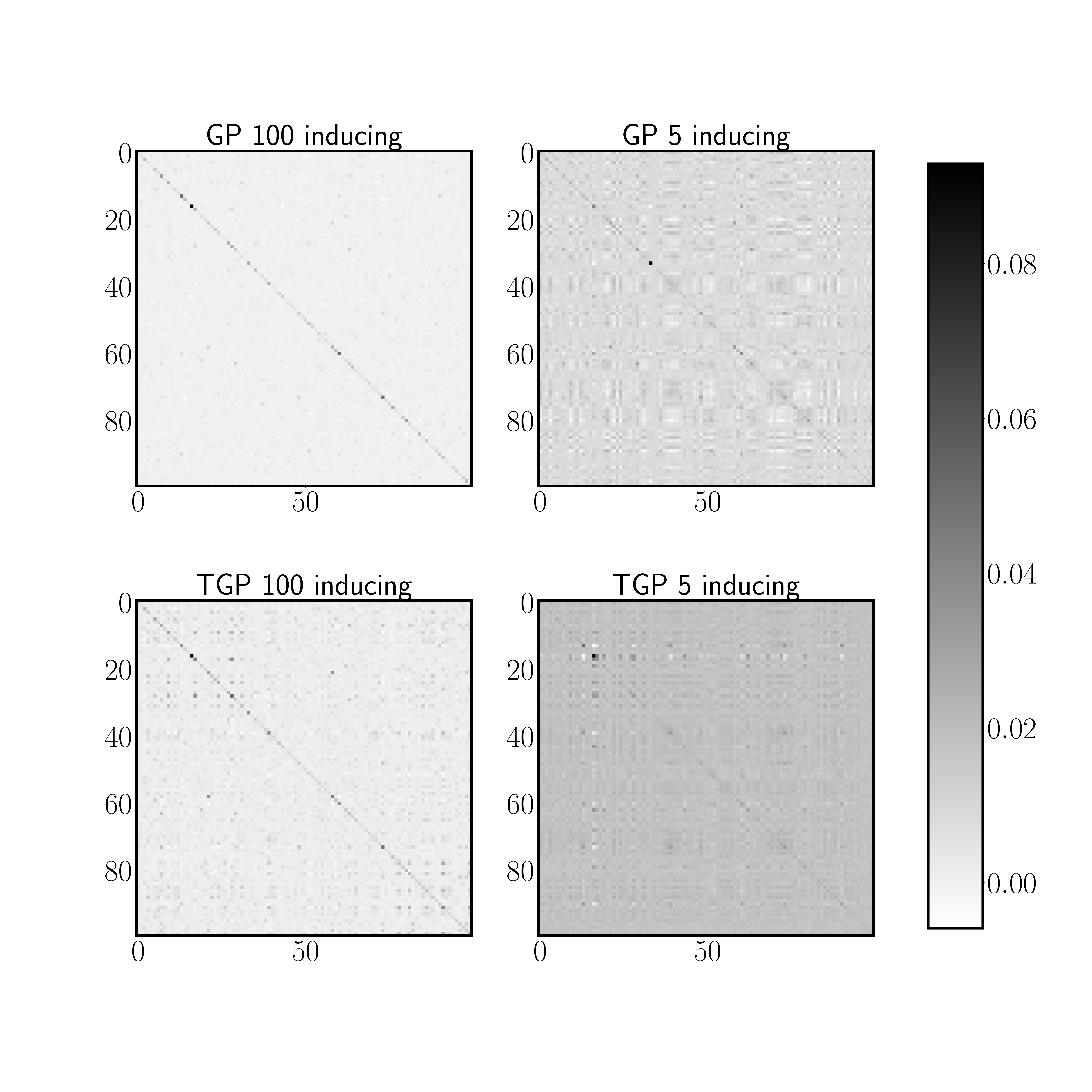}
    \caption{Covariance from $q(\Tfinit)$ evaluated at 100 training points}
\end{subfigure}
\begin{subfigure}[t]{0.5\textwidth}
    \centering
    \includegraphics[width=\textwidth]{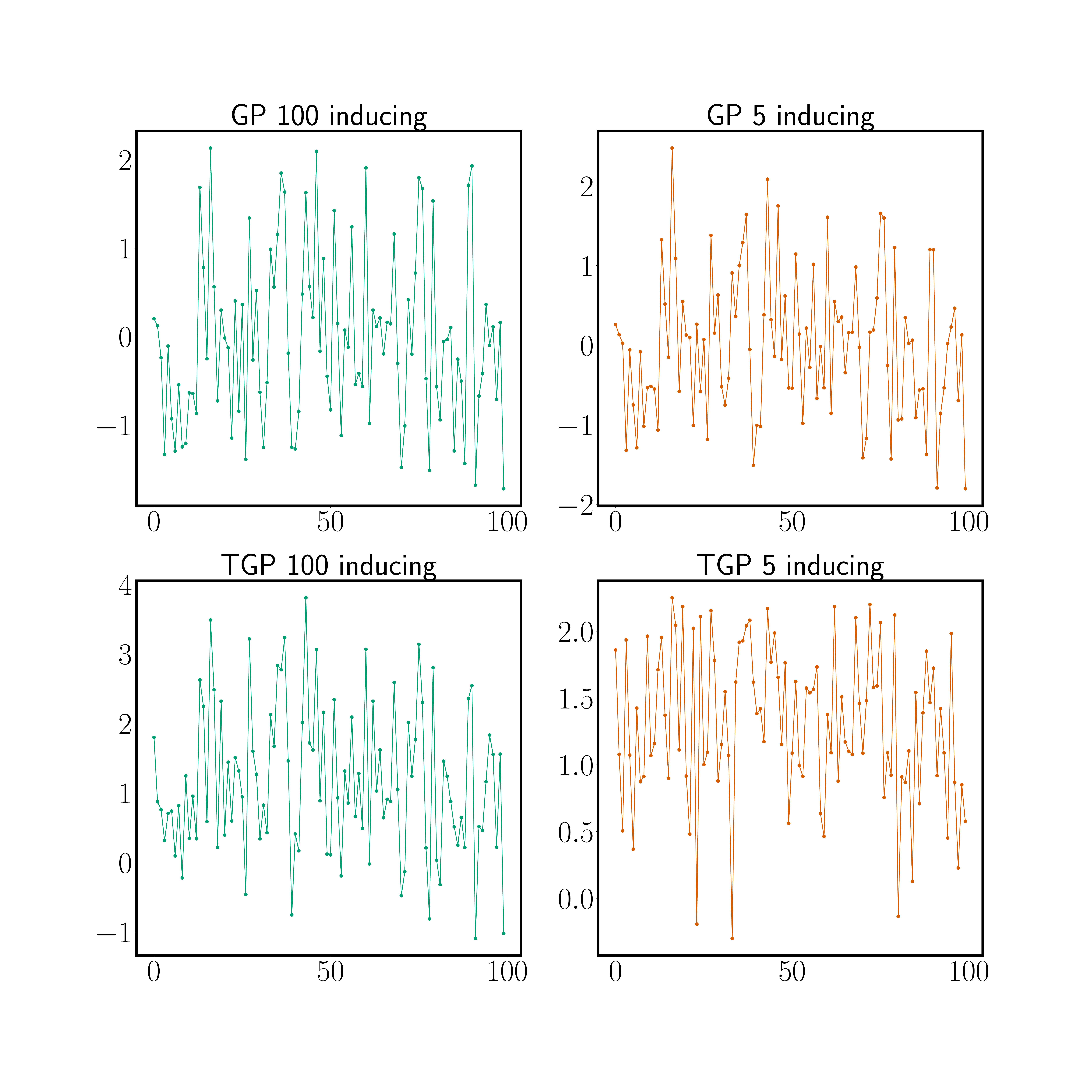}
    \caption{Mean from $q(\Tfinit)$ evaluated at 100 training points}
\end{subfigure}
\caption{This figure shows the mean and covariance from the \GP variational distribution $q(\Tfinit)$ evaluated at 100 training points. As shown in the plot the covariance have not collapse to a point mass (i.e the cells would have to be completely white) and the mean also change across training points. This means that the model has not learned to just output a constant value for $\Tfinit$ and model everything through the Neural Network.}
\label{fig:cov_mean_qf0}
\end{figure}

To do so, we pick the $\texttt{concrete}$ dataset, which is one of the datasets in which this effect is presented. For this dataset we plot the mean and  covariance of $q(\Tfinit)$ at 100 random training locations in \fig \ref{fig:cov_mean_qf0}. As we see in the plot, the mean and covariance from $q(\Tfinit)$ has not collapsed to a constant distribution. This means that the model has not learn to output a constant value for $\Tfinit$ and perform all the modelling through the input-dependent $\theta(\W,\X^{(n)})$ model. Note however that understanding the specific role of the \GP and the \BNN in our model in terms of uncertainty quantification is something that we leave for future work.

}
\fi

\end{document}